\documentclass[lettersize,journal]{IEEEtran}
\usepackage{amsmath,amsfonts}
\usepackage{algorithmic}
\usepackage{algorithm}
\usepackage{array}
\usepackage[caption=false,font=normalsize,labelfont=sf,textfont=sf]{subfig}
\usepackage{textcomp}
\usepackage{stfloats}
\usepackage{url}
\usepackage{verbatim}
\usepackage{graphicx}
\usepackage{cite}
\usepackage{booktabs}
\usepackage{multirow} 
\usepackage{pifont}

\begin{document}

\title{ProSplat: Improved Feed-Forward 3D Gaussian Splatting for Wide-Baseline Sparse Views}

\author{
	Xiaohan~Lu,
        Jiaye~Fu,
	Jiaqi~Zhang,
        Zetian~Song,\\
        Chuanmin~Jia, ~\IEEEmembership{Member,~IEEE,}
	Siwei~Ma, ~\IEEEmembership{Fellow,~IEEE}\\	
		\thanks{
		}
		\thanks{Xiaohan~Lu, Jiaqi~Zhang, Zetian~Song, and Siwei~Ma are with the State Key Laboratory of Multimedia Information Processing, School of Computer Science, Peking University, Beijing 100871, China~(e-mail: luxiaohan@stu.pku.edu.cn, jqzhang@pku.edu.cn, ztsong@pku.edu.cn, swma@pku.edu.cn). }
        \thanks{Jiaye~Fu is with the State Key Laboratory of Multimedia Information Processing, School of Electronic and Computer Engineering, Peking University, Shenzhen, 518055, China, and also with the National Engineering Research Center of Visual Technology, Peking University, Beijing 100871, China (email: jyfu@stu.pku.edu.cn).}
        \thanks{Chuanmin~Jia is with the Wangxuan Institute of Computer Technology, State Key Laboratory of Multimedia Information Processing, Peking University, Beijing 100871, China~(email: cmjia@pku.edu.cn).}
	}
\maketitle
\markboth{IEEE Transactions on Circuits and Systems for Video Technology}%
{Shell \MakeLowercase{\textit{et al.}}: A Sample Article Using IEEEtran.cls for IEEE Journals}

\begin{abstract}
Feed-forward 3D Gaussian Splatting (3DGS) has recently demonstrated promising results for novel view synthesis (NVS) from sparse input views, particularly under narrow-baseline conditions. However, its performance significantly degrades in wide-baseline scenarios due to limited texture details and geometric inconsistencies across views.
To address these challenges, in this paper, we propose ProSplat, a two-stage feed-forward framework designed for high-fidelity rendering under wide-baseline conditions. The first stage involves generating 3D Gaussian primitives via a 3DGS generator. In the second stage, rendered views from these primitives are enhanced through an improvement model. Specifically, this improvement model is based on a one-step diffusion model, further optimized by our proposed Maximum Overlap Reference view Injection (MORI) and Distance-Weighted Epipolar Attention (DWEA). MORI supplements missing texture and color by strategically selecting a reference view with maximum viewpoint overlap, while DWEA enforces geometric consistency using epipolar constraints.
Additionally, we introduce a divide-and-conquer training strategy that aligns data distributions between the two stages through joint optimization. We evaluate ProSplat on the RealEstate10K and DL3DV-10K datasets under wide-baseline settings. Experimental results demonstrate that ProSplat achieves an average improvement of 1\,dB in PSNR compared to recent SOTA methods.

\end{abstract}

\section{Introduction}
Novel view synthesis (NVS) provides an efficient and flexible solution for reconstructing 3D scenes. However, synthesizing high-quality images from very sparse multi-view inputs, especially under wide-baseline settings, remains challenging due to limited texture information and difficulties in preserving geometric consistency across views. Both 3D scene reconstruction and NVS are critical to Free Viewpoint Video (FVV)\cite{gao2021dynamic}, Virtual Reality (VR)\cite{attal2020matryodshka}, and Augmented Reality (AR)~\cite{jin2025rendering}. NVS methods typically reconstruct a 3D scene from input views and subsequently render unseen viewpoints.
In this context, Neural Radiance Fields (NeRF)-based methods~\cite{mildenhall2021nerf,muller2022instant,fridovich2022plenoxels}, leveraging implicit representations, have recently achieved remarkable success. NeRF~\cite{mildenhall2021nerf} employs multilayer perceptrons (MLPs) to represent the radiance field of a scene and implements differentiable volume rendering to synthesize novel views. Despite their photorealistic rendering capabilities, these implicit methods incur substantial computational costs, limiting their practicality for real-time applications. Alternatively, 3D Gaussian Splatting (3DGS)\cite{kerbl20233d3dgs} has emerged as a promising explicit representation, using rasterization-based rendering\cite{kopanas2022neural} to balance efficiency and view quality.

However, conventional 3DGS suffers from two primary drawbacks. First, it relies on per-scene parameter optimization, which takes several minutes and slows the 3D reconstruction process. Second, 3DGS requires densely captured input views to achieve high visual fidelity. To overcome these limitations, feed-forward methods have emerged, such as pixelSplat~\cite{pixelsplat}, MVSplat~\cite{mvsplat}, and DepthSplat~\cite{xu2024depthsplat}, which reconstruct 3D scenes using only a feed-forward network, eliminating per-scene training. These methods train a general generative model on large datasets and perform inference to estimate the parameters of 3D Gaussian primitives. They effectively utilize sparse input views, addressing the key shortcomings of optimized/vanilla 3DGS and motivating our focus on feed-forward methods.

Currently, most existing feed-forward methods~\cite{pixelsplat,mvsplat,xu2024depthsplat,zhang2024gaussian,yu2021pixelnerf} focus on narrow-baseline scenarios in which the overlap between adjacent views is substantial. Although these methods perform well under such conditions, narrow-baseline configurations limit view diversity, constraining the effectiveness of NVS applications. Extending high-quality reconstruction to wide-baseline scenarios remains a significant challenge. To tackle this challenge, several optimized 3DGS-based methods~\cite{yu2024viewcrafter,liu20243dgs,wu2025difix3d} have integrated video diffusion models~\cite{blattmann2023stable,xing2024dynamicrafter,rombach2022ldm} to generate higher-quality views, subsequently feeding these views back for iterative training. Among feed-forward methods, MVSplat360 \cite{chen2024mvsplat360} integrates a Latent Diffusion Model (LDM) \cite{rombach2022ldm} into its pipeline to refine the fidelity of rendered views. Nevertheless, the multi-step denoising and inter-frame attention mechanisms inherent to video diffusion models impose significant computational overhead, compromising the efficiency benefits of feed-forward methods. Additionally, MVSplat360 feeds the latent features of rendered views to LDM, failing to exploit pixel-level structure information rendered by 3DGS.
Recently, DIFIX3D+\cite{wu2025difix3d} employs a one-step image diffusion model\cite{parmar2024img2img,yin2024one} within an optimized 3DGS framework, achieving efficient high-quality rendering. However, it remains constrained by per-scene training requirements. 

To address these challenges, we propose \textbf{ProSplat}, a novel two-stage feed-forward framework that efficiently generates high-fidelity views under wide-baseline conditions without per-scene optimization.
In the first stage, ProSplat employs an efficient 3DGS generator, DepthSplat~\cite{xu2024depthsplat}, to produce 3D Gaussian primitives and render views from novel perspectives. In the second stage, these initial low-fidelity rendered views are enhanced using an improvement model based on a one-step diffusion model~\cite{parmar2024img2img}. This improvement model directly leverages the pixel-level structure information rendered by 3DGS and the rich 2D priors encoded in a pre-trained diffusion model (SD-Turbo~\cite{sauer2024adversarial_sd-turbo}), fine-tuned using LoRA~\cite{hu2022lora} for enhanced adaptation.

Specifically, we introduce Maximum Overlap Reference view Injection (MORI) and Distance-Weighted Epipolar Attention (DWEA) to enhance the geometric consistency and visual quality of the rendered views. MORI selects the most relevant input view as the reference view to supplement missing texture and color. DWEA effectively fuses geometrically corresponding regions of reference and rendered views within the latent space, reinforcing cross-view geometric consistency. The latent features obtained by downsampling with a Variational Autoencoder (VAE)\cite{kingma2013auto} encoder and U-Net\cite{ronneberger2015u} ensure computational efficiency and mitigate inconsistencies across network layers by confining feature fusion to localized regions. In addition, we further propose a divide-and-conquer training strategy, initially training the improvement model independently before integrating it into the feed-forward framework for joint optimization. This strategy aligns the output distribution of the 3DGS generator with the input requirements of the improvement model.

We evaluate ProSplat on benchmarks of large-scale datasets, DL3DV-10K~\cite{ling2024dl3dv} and RealEstate10K~\cite{zhou2018re10k}. Experimental results demonstrate that ProSplat consistently surpasses recent state-of-the-art (SOTA) methods across widely adopted metrics, achieving an average improvement of 1\,dB in PSNR. Our contributions are summarized as follows:
\begin{itemize}
\item We propose ProSplat, a two-stage feed-forward framework capable of synthesizing high-fidelity novel views under challenging wide-baseline conditions.
\item We introduce MORI, an effective strategy for selecting and injecting the most relevant input views to enhance texture and color fidelity.
\item We develop DWEA, a novel attention mechanism that leverages epipolar geometry to reinforce geometric consistency in latent feature representations.
\item We demonstrate the effectiveness and robustness of ProSplat through extensive experiments, consistently outperforming recent SOTA methods in terms of PSNR, SSIM~\cite{wang2004ssim}, and LPIPS~\cite{zhang2018lpips}.
\end{itemize}

The remainder of this paper is structured as follows. Section~\ref{sec:related works} reviews related works on sparse-view NVS, feed-forward 3DGS, and diffusion models for NVS. Section~\ref{sec:method} describes the ProSplat framework in detail. Section~\ref{sec:exeriment} presents experimental evaluations and demonstrates the advantages of ProSplat. Finally, Section~\ref{sec:conclusion} concludes the paper, while Section~\ref{sec:limitation} outlines limitations and future research directions.

\begin{figure*}[t]
\centering
\includegraphics[width=\linewidth, trim=0cm 16.9cm 0cm 0cm, clip]{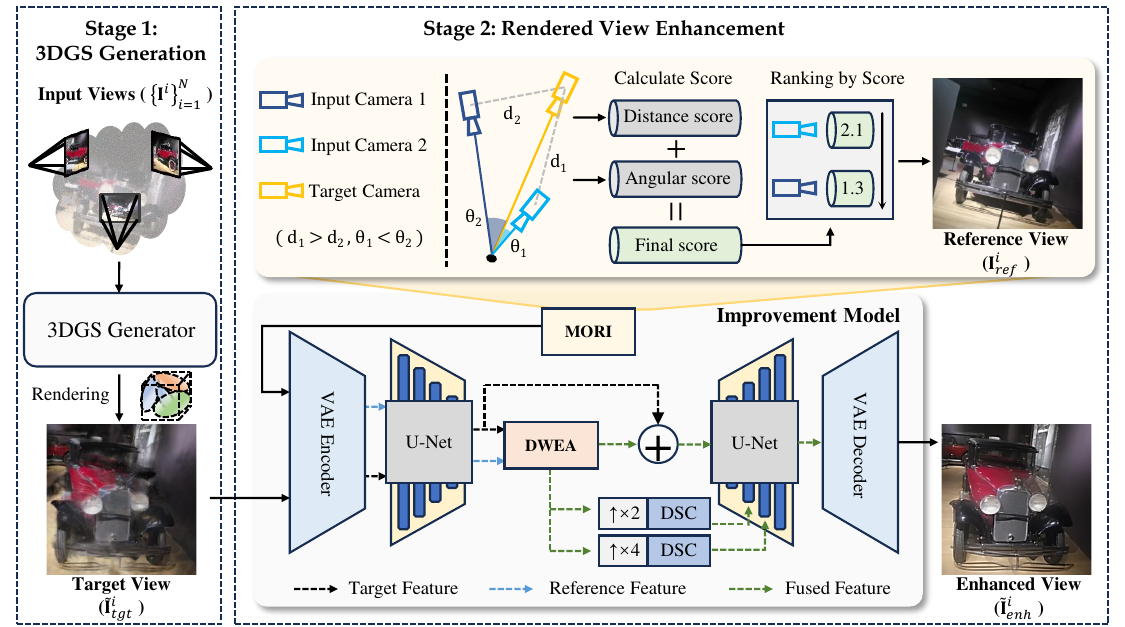}
\caption{\textbf{The overall framework of ProSplat.} The input views are fed into the 3DGS generator to obtain 3D Gaussian primitives and render target views. These views are enhanced by an improvement model (on the right side). The improvement model first uses MORI to select the most relevant input view as the reference view, based on the distance scores and angular scores. The reference and target views are encoded by VAE Encoder and U-Net to obtain the downsampled features. After these features are fused by DWEA, The fused feature is then upsampled and refined using Depthwise Separable Convolutions (DSC)\cite{chollet2017xception}, and injected into the expansive path of the U-Net. Finally, the enhanced view is reconstructed via the VAE decoder.}
\label{framework}
\end{figure*}
\section{Related Works}
\label{sec:related works}
\subsection{Sparse-View NVS}
Novel view synthesis aims to generate photorealistic images of viewpoints that have not been directly observed.
Early works~\cite{chen2023nvs1,gortler2023nvs2} primarily rely on geometric methods, such as ray interpolation over densely captured image sets. However, these methods required extensive input views, limiting their practicality in sparse-view scenarios.
With the advent of deep learning, learning-based methods~\cite{flynn2016deepstereonvs,kalantari2016learningnvs,wang2024dust3r} have emerged, enabling inference of scene geometry and appearance from sparse inputs. These methods offer greater flexibility and enhanced data efficiency.
More recently, methods~\cite{muller2022instant,fridovich2022plenoxels} based on NeRF~\cite{mildenhall2021nerf} have leveraged substantial MLP backbones and dense volumetric sampling to achieve high-quality rendering. Nevertheless, the reliance on implicit representations introduces significant computational overhead, constraining their use in time-sensitive applications.

Unlike implicit methods, 3DGS~\cite{kerbl20233d3dgs} provides an explicit representation, delivering superior quality with real-time rendering. However, both 3DGS and NeRF require dense input views for achieving high visual fidelity. To mitigate this limitation, several works introduce geometric prior regularization~\cite{niemeyer2022regnerf,somraj2023simplenerf} or leverage pre-trained models to estimate depth ~\cite{deng2022depthnerf,zhu2024fsgs} and normal ~\cite{yu2022monosdnerf} as supervision. DropGaussian\cite{park2025dropgaussian} randomly removes Gaussian primitives during training to increase the updating opportunities and improve visibility for the remaining ones under sparse-view conditions. Ma \textit{et al.}\cite{ma2025novel} propose an attention-based illumination model that exploits light fields from neighboring views. Although these methods achieve good performance for views near the inputs, they typically produce artifacts such as blurring or noise when rendering views that are farther away.
\subsection{Feed-Forward 3DGS}
Vanilla 3DGS relies on per-scene optimization, which limits its practicality due to its high computational cost.
To address this limitation, several feed-forward methods ~\cite{zhang2024gs,tang2024lgm,szymanowicz2024splatter,zheng2024gps,zhou2025gps} have been developed for rapid 3D reconstruction. Some methods ~\cite{zheng2024gps,zhou2025gps} generate 3D Gaussian representations of human bodies, while others~\cite{zhang2024gs,tang2024lgm,szymanowicz2024splatter} focus on generating 3D assets. These object-level scenes have well-defined boundaries and benefit from strong structural priors, facilitating accurate and high-fidelity reconstruction.
However, scene-level reconstruction remains inherently challenging due to complex layouts and limited structural priors when only sparse input views are available. Recently, some scene-level methods~\cite{pixelsplat,mvsplat,xu2024depthsplat,noposplat,wewer2024latentsplat} have been proposed. PixelSplat~\cite{pixelsplat} regresses pixel-aligned 3D Gaussian primitives from image pairs by predicting a probabilistic depth distribution for each input view. However, accurately estimating depth distributions solely from image features extracted by an epipolar transformer is challenging, often resulting in noisy and low-quality geometric reconstructions. To overcome these limitations, MVSplat~\cite{mvsplat} introduces cost volumes that encode cross-view feature similarities across depth candidates, providing robust geometric cues for localizing 3D Gaussian primitives. DepthSplat\cite{xu2024depthsplat} further integrates monocular features from a pre-trained monocular depth network, significantly enhancing robustness in challenging scenarios like low-textured regions and reflective surfaces.

However, these scene-level methods typically address narrow-baseline scenarios, which are characterized by small disparities across views. This restricts their ability to generalize to wide-baseline settings. To tackle this limitation, MVSplat360\cite{chen2024mvsplat360} targets wide-baseline conditions and demonstrates the capability in reconstructing full $360^{\circ}$ scenes from only five input views. MVSplat360 integrates a latent diffusion model~\cite{rombach2022ldm} into the feed-forward pipeline to enhance rendering quality using latent features of the rendered views. Nevertheless, video diffusion models involve multiple denoising steps and inter-frame attention, significantly reducing their efficiency and suitability for real-time applications. Moreover, MVSplat360 relies on latent features of rendered views as input to the LDM, failing to exploit pixel-level structure information rendered by 3DGS. In contrast, our proposed approach uses a one-step diffusion model directly on rendered images, achieving superior novel view synthesis performance while maintaining feed-forward efficiency.

\subsection{Diffusion Models for NVS}
Large-scale diffusion models have recently shown impressive capabilities in synthesizing realistic content by leveraging substantial priors from extensive training datasets. These models are particularly well-suited for novel view synthesis tasks. Methods such as Zero-1-to-3\cite{liu2023zero}, Zero123Plus\cite{shi2023zero123++}, and MVDream\cite{shi2023mvdream} generate multi-view images by fine-tuning large-scale pre-trained diffusion models, employing cross-view attention mechanisms for implicit 3D consistency. Building on this direction, EpiDiff\cite{huang2024epidiff} introduces epipolar attention explicitly enforcing 3D constraints, improving cross-view consistency. SparseFusion\cite{zhou2023sparsefusion} utilizes a distilled latent video diffusion model~\cite{rombach2022ldm} to directly recover a plausible 3D representation of 3D objects. 
Recently, several approaches have emerged that integrate diffusion models with 3DGS\cite{kerbl20233d3dgs} and NeRF\cite{mildenhall2021nerf} for NVS.
Viewcrafter\cite{yu2024viewcrafter}, 3D-Enhancer\cite{liu20243dgs}, and GenFusion\cite{wu2025genfusion} incorporate video diffusion models~\cite{blattmann2023stable,xing2024dynamicrafter,rombach2022ldm} for high-fidelity rendering, but these methods require per-scene training. 
Feed-forward methods such as MVSplat360\cite{chen2024mvsplat360} mitigate per-scene training by using latent diffusion model\cite{rombach2022ldm}. More recently, methods including DIFIX3D+\cite{wu2025difix3d}, NerDiff\cite{gu2023nerfdiff}, and GeNVS\cite{chan2023generative} employ image diffusion models for rapid enhancement of rendered views. While NerDiff and GeNVS are designed for NeRF-like methods, DIFIX3D+ is tailored for optimized 3DGS which requires per-scene training. DIFIX3D+ uses a one-step diffusion model to enhance the rendered views, which are subsequently fed back into the training stage for an additional round of optimization. In contrast, ProSplat employs a one-step diffusion model yet adopts a purely feed-forward framework, eliminating per-scene training requirements and enabling high-quality, efficient novel view synthesis exclusively during rendering.
\section{Method}
\label{sec:method}

Given $N$ input sparse views $\mathcal{I} = \left\{ \mathbf{I}^i \right\}_{i=1}^{N}$, our goal is to predict per-pixel parameters for 3DGS and render novel photorealistic views. ProSplat is a two-stage framework: 3DGS generation and rendered view enhancement. The overall framework of ProSplat is shown in Figure~\ref{framework}. 
In the first stage, we generate 3D Gaussian primitives and render novel views. 
In the second stage, the rendered views are enhanced by a view improvement model. The improvement model is based on a one-step diffusion model\cite{parmar2024img2img} that has been carefully optimized. 
These two stages are closely interconnected, as the 3DGS generation stage estimates the 3D geometric structure, and the rendered view enhancement stage leverages 2D image priors from a pre-trained diffusion model to improve the visual fidelity of the rendered views.

Section~\ref{sec:preliminary} gives a brief introduction to 3DGS, as ProSplat benefits from its exceptional performance, including the concept of feed-forward 3DGS for efficient scene reconstruction. In Section~\ref{sec:3dgs_generator}, we introduce the 3DGS generator in the first stage of ProSplat. In Section~\ref{sec:improvement_model}, we provide a detailed description of the proposed improvement model in the second stage. In Section~\ref{sec:train_obj}, we present the training objectives and optimization strategies employed in ProSplat.
\subsection{Preliminary: 3DGS and Feed-Forward Generation}
\label{sec:preliminary}
3DGS reconstructs a static 3D scene explicitly with 3D Gaussian primitives that are defined by a 3D covariance matrix $\bf{\Sigma} \in \mathbb{R}^{3 \times 3}$ and location (mean) $\bf{\mu}\in \mathbb{R}^{3}$, 
\begin{equation}
G(\mathbf{x}) = \exp\left( -\frac{1}{2}(\mathbf{x} - \mathbf{\mu})^\top \Sigma^{-1} (\mathbf{x} - \mathbf{\mu}) \right),
\end{equation}
where $\mathbf{x}$ is the coordinate of an arbitrary point. The covariance matrix $\mathbf{\Sigma}$ can be decomposed into a scaling matrix $\mathbf{S}$ and a rotation matrix $\mathbf{R}$ as 
\begin{equation}
\mathbf{\Sigma} = \mathbf{R} \mathbf{S} \mathbf{S}^\top \mathbf{R}^\top,
\end{equation}
where $\mathbf{S}$ is determined by scaling factor $\mathbf{s} \in \mathbb{R}^{3}$ and $\mathbf{R}$ is derived by a rotation quaternion $\mathbf{q} \in \mathbb{R}^{4}$.

For rendering, all 3D Gaussian primitives are projected onto 2D planes using a differentiable Gaussian splatting pipeline\cite{zwicker2001surface}. First, covariance matrix $\mathbf{\Sigma'}$ in camera coordinates is computed as
\begin{equation}
\mathbf{\Sigma'} = \mathbf{J} \mathbf{W} \mathbf{\Sigma} \mathbf{W}^\top \mathbf{J}^\top,
\end{equation}
where $\mathbf{W}$ represents the viewing transformation matrix and $\mathbf{J}$ denotes the Jacobian matrix of the affine approximation of the projective transformation. Next, a differentiable splatting method is employed to project the 3D Gaussian spheres onto 2D Gaussian distributions, ensuring efficient $\alpha$-blending for rendering and color supervision. The color is represented by spherical harmonics (SH) coefficients, which are associated with the camera poses. For each pixel, the color is rendered using sorted 3D Gaussian primitives based on the distance between the pixel and the center of the 3DGS primitives as 
\begin{equation}
C = \sum_{i \in M} c_i \alpha_i \prod_{j=1}^{i-1} (1 - \alpha_j),
\end{equation}
where $c_i$ is the color derived by SH coefficients and $\alpha_i$ is opacity.

To summarize, the original 3DGS model characterizes each Gaussian primitive by the following attributes: a 3D position $\mathbf{\mu}$, a color defined by $SH$ coefficients, a rotation represented by a quaternion $\mathbf{r}$, a scaling factor $\mathbf{s}$, and an opacity $\alpha$. For feed-forward 3DGS generation from sparse views, the goal is to obtain a 3DGS reconstruction of scenes from each image: 
\begin{equation}
f_\theta : \left\{ \left( \mathbf{I}^i, \mathbf{P}^i \right) \right\}_{i=1}^{M} \rightarrow \left\{ \left( \mathbf{\mu}_j, \mathbf{SH}_j, \mathbf{r}_j, \mathbf{s}_j, \alpha_j \right) \right\}_{j=1}^{H \times W \times M},
\end{equation}
where $f_\theta$ is a feed-forward network and $\theta$ are the learnable parameters optimized from a large-scale training dataset.
Additionally, $\mathbf{I}^i$ denotes an input image with the corresponding camera projection matrices $\mathbf{P}^i$, and $M$ represents the number of such images. $H$ and $W$ are the height and width of the images, respectively. The total number of Gaussian primitives corresponds to the combined pixel count across all input views.
\subsection{3DGS Generation}
\label{sec:3dgs_generator}
ProSplat uses DepthSplat\cite{xu2024depthsplat}, a feed-forward 3DGS generator, to generate 3D Gaussian primitives. The generator concatenates multi-view features by constructing cost volumes and monocular features from a pre-trained monocular depth network, which are then used to generate 3D Gaussian primitives. Specifically, for cost volumes, convolutions are employed to extract multi-scale features from each input view. Then, a multi-view Swin Transformer~\cite{liu2021swin} is applied to obtain multi-view features $\{ \mathbf{F}^i \}_{i=1}^{N}$. Next, $D$ depth candidates $\{ d_m \}_{m=1}^{D}$ are uniformly sampled from the near and far ranges. The feature from view $j$ is then warped to view $i$ to obtain a warped feature $\{ \mathbf{F}^{j \to i}_{d_m} \}_{m=1}^{D}$, using the camera projection matrix and each depth candidate $d_m$. The cost volume $\{ \mathbf{C}^i \}_{i=1}^{N}$ for view $i$ is obtained by performing a dot-product operation between the warped feature $\{ \mathbf{F}^{j \to i}_{d_m} \}_{m=1}^{D}$ and the feature $\{ \mathbf{F}^i \}_{i=1}^{N}$.

For monocular features, a Vision Transformer (ViT) model is employed to extract monocular features for each input view. These features have the same resolution as the cost volume, represented as $\{\mathbf{F}^i_{\text{mo}}\}_{i=1}^{N}$. Both the cost volume and the monocular features are obtained for each view, represented as $\{ \mathbf{C}^i \}_{i=1}^{N}$ and $\{\mathbf{F}^i_{\text{mo}}\}_{i=1}^{N}$, respectively. These features are concatenated and then processed by a 2D U-Net to regress depth maps for generating 3D Gaussian primitives. Finally, these primitives are passed through an MLP to render novel views $\tilde{\mathbf{I}}^i_{tgt}$.

\subsection{Rendered View Enhancement}
\label{sec:improvement_model}
As illustrated in the right region of Figure~\ref{framework}, the $i$-th rendered view (the target view) $\tilde{\mathbf{I}}^i_{tgt}$ and the input views $\left\{ \mathbf{I}^i \right\}_{i=1}^{N}$, together with the corresponding camera poses $ \mathbf{P}^i_{tgt} $ and $\left\{\mathbf{P}^i\right\}_{i=1}^{N}$, respectively, are fed into the improvement model. 
First, a reference view $\mathbf{I}^i_{\text{ref}}$ and its corresponding camera pose $\mathbf{P}^i_{\text{ref}}$ are selected from input views using the proposed MORI strategy. This reference view provides supplementary detailed texture and color information required by the target view. 
The VAE encoder then encodes both the target view and reference view in latent space, enriching the representational capacity of the diffusion model. Second, the latent features are fed into a U-Net to predict the noise: 
\begin{equation}
\mathbf{\epsilon}^i = \mathcal{U}_\theta(\tilde{\mathbf{z}}^i_{int}, \mathbf{z}_{ref}^i, \mathbf{P}^i_{tgt},\mathbf{P}_{ref}^i, t),
\end{equation}
where t is timestep and $\mathcal{U}_\theta$ denotes the U-Net with $\theta$ being its trainable parameters. $\tilde{\mathbf{z}}^i_{int}$ and $ \mathbf{z}_{ref}^i$ represent latent features of the target view and reference view, respectively. Within the U-Net architecture, we apply the proposed DWEA to the bottleneck features, and propagate the resulting attention-enhanced representations through the expansive path to guide the subsequent decoding process. DWEA enhances geometric consistency across views in combination with MORI. Subsequently, the noise predicted by the U-Net is used for the one-step denoising process to generate the enhanced latent feature $\mathbf{\tilde{\mathbf{z}}}^i_{enh}$:
\begin{equation}
\mathbf{\tilde{\bf{z}}}^i_{enh} = \mathcal{D}(\tilde{\bf{z}}^i_{int},\epsilon^i,t).
\end{equation}
Finally, the VAE decoder decodes the latent feature $\mathbf{\tilde{\bf{z}}}^i_{enh}$ back into a high-fidelity image $\mathbf{\tilde{\bf{I}}}^i_{enh}$.

In the following subsections, we present the details of MORI and DWEA. 
\begin{figure}[t]
\centering
\includegraphics[width=\linewidth, trim=0cm 22.2cm 10cm 0cm, clip]{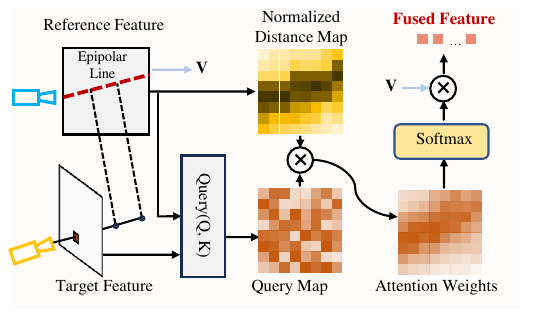}
\caption{\textbf{The flow of our Distance-Weighted Epipolar Attention.} Target and reference features denote the feature maps extracted from the target and reference views, respectively. For each feature vector in the target feature map, a query operation is performed over the reference feature map to generate a query map. This query map is then element-wise multiplied by the corresponding epipolar distance map. The result is passed through a sigmoid activation, and subsequently multiplied by the reference feature’s value vector $\mathbf{V}$, producing the final fused feature.}
\label{epipolar_attn}
\end{figure}

\subsubsection{Maximum Overlap Reference view Injection}
\label{subsec:reference_inject}
The rendered view (target view) cannot be effectively enhanced without incorporating a reference view, due to the lack of texture and color information. Therefore, we propose MORI that selects the input view with the highest overlap with the target view, ensuring that the most relevant information is utilized. MORI reduces the influence of irrelevant content introduced by input views with lower overlap, thus improving the reliability of the enhancement process.

To compute the overlap, we evaluate the similarity between each target view $\tilde{\mathbf{I}}^i_{tgt}$ and all input views $\left\{ \mathbf{I}^i \right\}_{i=1}^{N}$. First, we calculate the Euclidean distance between the translation vectors of the target view $\mathbf{T}^i_{tgt}$ and an input view $\mathbf{T}^j$:
\begin{equation}
\mathbf{Dist}^{i,j}_{tgt} = \Vert \mathbf{T}^i_{tgt} - \mathbf{T}^j\Vert,
\end{equation}
where $\mathbf{Dist}^{i,j}_{tgt}$ represents the distance score between two views. Although this distance serves as an indication of view overlap, it may not capture alignment in certain cases. As illustrated in the top-right region of Figure~\ref{framework}, while \textit{Input Camera 2} is spatially close to \textit{Target Camera}, its viewing direction diverges significantly. In contrast, \textit{Input Camera 1} exhibits higher directional alignment with \textit{Target Camera}. Therefore, we additionally incorporate a measure of angular similarity to better reflect view overlap. More precisely, we use the third column of the rotation matrices, which represents the viewing direction along the z-axis, denoted as $\mathbf{R}^i_{tgt}(z)$ and $\mathbf{R}^i(z)$, respectively. The angular score is computed as:

\begin{equation}
\mathbf{Angle}^{i,j} = \frac{\mathbf{R}^i_{tgt}(z)}{\| \mathbf{R}^i_{tgt}(z) \|} \cdot \frac{\mathbf{R}^j(z)}{\| \mathbf{R}^j(z) \|},
\end{equation}
where $\mathbf{Angle}^{i,j}\in[-1,1]$ denotes the cosine similarity. We then compute the final overlap score between the target view $i$ and input view $j$ as:
\begin{equation}
\mathbf{Score}^{i,j} = \frac{1}{\mathbf{Dist}^{i,j}_{tgt}} + \frac{1}{2}(\mathbf{Angle}^{i,j}+1), 
\end{equation}
where the angular score is normalized to the range $[0,1]$ for compatibility with the distance score.
The input view with the highest overlap score is selected as the reference view. This reference is encoded by the VAE and injected into the expansive path of the U-Net to extract intermediate features, which are subsequently fused with the feature of the target view to form a combined representation for enhancement.

\subsubsection{Distance-Weighted Epipolar Attention}
\label{subsec:epipolar_atten}
We propose an attention mechanism, DWEA, to enhance the cross-view feature matching between the target and reference views. As shown in Figure~\ref{epipolar_attn}, DWEA fuses the feature of the target view with the corresponding feature of the reference view. Although global dot-product attention \cite{vaswani2017attention} can model long-range dependencies, it may result in imprecise correspondences between different views due to the lack of geometric constraints. To address this limitation, DWEA modulates global attention weights based on the epipolar geometry between views, thereby improving spatial consistency. More precisely, we introduce a modulation factor derived from the geometric relationship between a pixel and its corresponding epipolar line in the reference view. This factor is then multiplied with the global attention weights, which are computed via dot products between the query and key feature vectors, to guide the attention mechanism using geometric constraints.

Let $\mathbf{R}^i_{tgt}$ and $\mathbf{T}^i_{tgt}$ denote the rotation matrix and the translation vector for the target view, and $\mathbf{R}^i_{ref}$ and $\mathbf{T}^i_{ref}$ denote the corresponding parameters of the reference view. For clarity, we omit the view index $i$ throughout this section.
The coordinate of a pixel in the target view is first extended from 2D to homogeneous 3D space:
\begin{equation}
\mathbf{coord} = [x,y,1]^\top,
\end{equation}
where $x$ and $y$ are the horizontal and vertical pixel coordinates, respectively. Next, we compute the relative translation from the target view to the reference view:
\begin{equation}
\mathbf{t} = \mathbf{T}_{ref} - \mathbf{R}_{tgt}^\top \cdot \mathbf{T}_{tgt}, 
\end{equation}
\begin{equation}
\mathbf{t}_m = \begin{bmatrix}
0 & -t_z & t_y \\
t_z & 0 & -t_x \\
-t_y & t_x & 0
\end{bmatrix},
\end{equation}
where $t_x$, $t_y$, and $t_z$ are the three components of $\mathbf{t}$. Using these transformations, the fundamental matrix $\mathbf{F}$ is computed to derive the epipolar line:
\begin{equation}
\mathbf{F} = \mathbf{K}_{ref}^{-1} \cdot {\mathbf{t}_m} \cdot \mathbf{R}_{ref} \cdot \mathbf{R}_{tgt}^{\top} \cdot \mathbf{K}_{tgt}^{-1}, 
\end{equation}
where $\mathbf{K}_{tgt}$ and $\mathbf{K}_{ref}$ are the intrinsic matrices of the target and reference views, respectively. The parameters of the epipolar line are then derived as 
\begin{equation}
[a, b, c] = \mathbf{F} \cdot \mathbf{coord}, 
\end{equation}
which corresponds to the homogeneous representation of the line equation $ax+by+c=0$.
The distance between the pixel and its corresponding epipolar line is computed to yield a distance map $\mathbf{d}$. This distance map serves as a modulation factor for the global attention weights:
\begin{equation}
\mathbf{attn}_{comb} = \mathbf{attn}_{g} \cdot \operatorname{Norm}(\exp(-\mathbf{d})), 
\end{equation}
where $\mathbf{attn}_{g}$ denotes the original dot-product attention scores and $\operatorname{Norm}(\cdot)$ indicates min-max normalization to the range $[0,1]$. 
Finally, the adjusted attention map is passed through a sigmoid function and multiplied by the value vectors from the reference view to obtain the fused feature representation.

This fused feature is then added to the target feature and passed into the expansive path of the U-Net. To maintain consistency across feature resolutions, the fused feature is upsampled by factors of $2\times$ and $4\times$, and processed using a Depthwise Separable Convolution~\cite{chollet2017xception}, consisting of a depthwise convolution followed by a pointwise convolution. The resulting features are then incorporated into subsequent stages of the expansive path to enhance spatial fidelity.
\subsection{Training Objectives}
\label{sec:train_obj}
\subsubsection{Divide-and-conquer Strategy}
We adopt a divide-and-conquer aggregation strategy to effectively train ProSplat. 
First, the 3DGS Generator is initialized with pre-trained parameters, while the improvement model is trained independently. After the independent training phase, the improvement model is integrated with the 3DGS Generator for joint training. 
During joint training, we freeze all components of the improvement model except for the VAE encoder. 
Simultaneously, within the 3DGS Generator, the monocular feature and multi-view feature extraction networks remain fixed, while only the 3D Gaussian adapter network is fine-tuned.
This strategy ensures that the output distributions of the 3DGS generator align appropriately with the requirements of the improvement model.
\subsubsection{Dataset Curation}
First, we use the 3DGS Generator to render views based on training data from the DL3DV-10K\cite{ling2024dl3dv} and RealEstate10K\cite{zhou2018re10k} datasets. 
These rendered views subsequently serve as target views for the improvement model. 
Concurrently, we identify and save the nearest reference views and the corresponding camera parameters, as detailed in Section~\ref{sec:improvement_model}. To manage the dataset size effectively, we select between 5 to 7 target views per scene. Finally, we store the ground truth images corresponding to each target view, thereby creating paired datasets derived from the original datasets for training the improvement model.
\subsubsection{Training Loss of improvement model}
We train the improvement model with a combination
of mean squared error (MSE) and LPIPS \cite{zhang2018lpips} losses computed between the colors of each enhanced view $\tilde{\mathbf{I}}^i_{enh}$ and ground truth view $\mathbf{I}_{\text{gt}}^i$:
\begin{equation}
\mathcal{L} = \text{MSE}(\tilde{\mathbf{I}}^i_{enh}, \mathbf{I}_{\text{gt}}^i) + \lambda \cdot \text{LPIPS}(\tilde{\mathbf{I}}^i_{enh}, \mathbf{I}_{\text{gt}}^i)
, 
\end{equation}
where $\lambda$ is empirically set to 5.
\subsubsection{Training Loss of joint training}
For the joint training, we calculate the total loss as the sum of MSE and LPIPS losses over all views generated in a single forward pass: 
\begin{equation}
\mathcal{L}_{joint} = \sum_{m=1}^{M} \left( \text{MSE}(\tilde{\mathbf{I}}^m_{enh}, \mathbf{I}_{\text{gt}}^m) + \lambda \cdot \text{LPIPS}(\tilde{\mathbf{I}}^m_{enh}, \mathbf{I}_{\text{gt}}^m) \right)
, 
\end{equation}
where M is the number of target views processed in each forward pass.
\begin{table*}[htbp]
  \centering
  \caption{\textbf{Comparison with State-of-the-Art Methods on DL3DV-10K Using All Scan-Round Views and ONLY THE FIRST SCAN ROUND.} The \textbf{views} in the table correspond to the number of input views.}
  \label{tab:dl3dv_n300}
  \renewcommand{\arraystretch}{1.3} 
  \setlength{\tabcolsep}{4pt}       
  \begin{tabular}{c|l|cccccccccccc} 
    \toprule
    \multirow{2}{*}{\textbf{Mode}} & \multirow{2}{*}{\textbf{Method}} 
    & \multicolumn{4}{c}{\textbf{4 views}} 
    & \multicolumn{4}{c}{\textbf{5 views}} 
    & \multicolumn{4}{c}{\textbf{6 views}} \\
    \cmidrule(lr){3-6} \cmidrule(lr){7-10} \cmidrule(lr){11-14}
     & & PSNR↑ & SSIM↑ & LPIPS↓ & FID↓ 
       & PSNR↑ & SSIM↑ & LPIPS↓ & FID↓ 
       & PSNR↑ & SSIM↑ & LPIPS↓ & FID↓ \\
    \midrule

    \multirow{5}{*}{All round}
     & pixelSplat\cite{pixelsplat}    & 14.34 & 0.342 & 0.613 & 169.03 & 15.63 & 0.406 & 0.566 & 153.94 & 15.49 & 0.394 & 0.554 & 144.18 \\
     & MVSplat\cite{mvsplat}      & 15.87 & 0.434 & 0.504 & 81.68  & 16.45 & 0.467 & 0.473 & 68.37  & 16.77 & 0.485 & 0.456 & 62.51  \\
     & MVSplat360\cite{chen2024mvsplat360}   & 15.91 & 0.464 & 0.448 & 18.79  & 16.81 & 0.514 & 0.418 & 17.01  & 17.09 & 0.514 & 0.418 & 16.54  \\
     & DepthSplat\cite{xu2024depthsplat}   & 16.67 & 0.494 & 0.422 & 44.37  & 17.47 & 0.537 & 0.387 & 34.98  & 17.87 & 0.559 & 0.371 & 32.09  \\
     & \textbf{ProSplat} (Ours) & \textbf{18.20} & \textbf{0.536} & \textbf{0.347} & \textbf{18.16} 
                    & \textbf{18.90} & \textbf{0.572} & \textbf{0.317} & \textbf{15.20} 
                    & \textbf{19.24} & \textbf{0.590} & \textbf{0.302} & \textbf{14.29} \\
    \midrule

    \multirow{5}{*}{First round}
     & pixelSplat\cite{pixelsplat}    & 15.29 & 0.367 & 0.585 & 145.34 & 16.60 & 0.440 & 0.539 & 138.70 & 16.95 & 0.493 & 0.501 & 101.35 \\
     & MVSplat\cite{mvsplat}      & 17.07 & 0.502 & 0.438 & 62.43  & 17.64 & 0.531 & 0.405 & 47.52  & 17.97 & 0.542 & 0.398 & 35.46  \\
     & MVSplat360\cite{chen2024mvsplat360}   & 17.25 & 0.517 & 0.417 & 20.16  & 17.81 & 0.562 & 0.352 & 18.89  & 18.04 & 0.590 & 0.342 & 17.45  \\
     & DepthSplat\cite{xu2024depthsplat}   & 18.27 & 0.579 & 0.348 & 33.64  & 19.10 & 0.619 & 0.314 & 26.23  & 19.70 & 0.653 & 0.290 & 22.93  \\
     & \textbf{ProSplat} (Ours)    & \textbf{19.83} & \textbf{0.610} & \textbf{0.285} & \textbf{16.54} 
                    & \textbf{20.28} & \textbf{0.635} & \textbf{0.264} & \textbf{14.73} 
                    & \textbf{21.12} & \textbf{0.668} & \textbf{0.240} & \textbf{13.08} \\
    \bottomrule
  \end{tabular}
\end{table*}

\section{Experimental Results}
\label{sec:exeriment}
\subsection{Experimental Details}
\subsubsection{Implementation details}
ProSplat is implemented using CUDA-based PyTorch\cite{paszke2019pytorch}, together with a CUDA-accelerated 3DGS renderer. For the 3DGS generator, we adopt DepthSplat\cite{xu2024depthsplat} and utilize its publicly released pretrained checkpoints. In the view enhancement stage, we employ SD-Turbo\cite{sauer2024adversarial_sd-turbo} and fine-tune it following the procedure of Pix2Pix-Turbo\cite{parmar2024img2img}, incorporating LoRA\cite{hu2022lora}. The text encoder is removed, and an empty prompt embedding is used, as semantic information is not required as a conditional input.
During training of the improvement model, we set a fixed learning rate of $1.5 \times 10^{-5}$ with a batch size of 2 for all datasets. 
During joint training, we freeze all components of the improvement model except for the VAE encoder and decoder, while in the 3DGS generator, only the Gaussian adapter network is fine-tuned. The joint model is trained with a learning rate of $1 \times 10^{-5}$ for 50,000 steps, using a batch size of 1.
\subsubsection{Datasets}
We evaluate ProSplat on two datasets: DL3DV-10K\cite{ling2024dl3dv} and RealEstate10K\cite{zhou2018re10k}, both of which feature wide-baseline view synthesis challenges. DL3DV-10K comprises 51.3 million frames captured from 10,510 real-world scenes, covering 65 point-of-interest categories. The dataset exhibits extensive view variation, with some scenes including full 360-degree camera transitions. We utilize the entire dataset for training, while the 140 benchmark scenes are excluded to prevent overfitting during evaluation. For each test scene, we follow the camera sampling strategy of MVSplat360\cite{chen2024mvsplat360}, selecting input views using farthest point sampling based on camera poses. We evaluate 56 target views per scene by uniformly sampling from the remaining views, resulting in a total of 7,840 test views.

The RealEstate10K\cite{zhou2018re10k} dataset consists of real estate videos sourced from YouTube. Diverse views of a scene are obtained by extracting frames from different timestamps, where each frame is associated with an estimated camera pose. 
To focus on the sparse-view reconstruction under wide-baseline conditions, we select scenes with large inter-view disparities and construct wide-baseline view pairs for testing. Specifically, we follow the NoPoSplat\cite{noposplat} methodology to compute the degree of view overlap scores using RoMa\cite{edstedt2024roma}, a SOTA feature matching algorithm. Reference and target views are defined as in NoPoSplat, and only scenes with an overlap score below 0.3 are selected.

\subsubsection{Metrics}
We conduct quantitative evaluation using PSNR, SSIM\cite{wang2004ssim}, and LPIPS\cite{zhang2018lpips}, compute the valid regions of the target views. In addition, following prior diffusion-based approaches, we adopt Fréchet Inception Distance (FID)\cite{heusel2017fid}  to assess image quality at the dataset level by computing the feature distributions of generated and real images.

\subsection{Results on the DL3DV-10K Dataset}
\subsubsection{Baselines}
We conduct a comprehensive comparison of ProSplat against recent SOTA feed-forward 3DGS models, including pixelSplat\cite{pixelsplat}, MVSplat\cite{mvsplat}, MVSplat360\cite{chen2024mvsplat360}, and DepthSplat\cite{xu2024depthsplat}. Among these baselines, MVSplat360 is explicitly tailored for wide-baseline inputs and therefore aligns with the objective of ProSplat. Although DepthSplat is not explicitly intended for this setting, it still achieves strong performance due to its use of monocular depth features. 
PixelSplat and MVSplat are high-performing generalizable 3DGS models that support multi-view inputs (more than two views) and are capable of reconstructing scenes with broader spatial coverage. We re-train MVSplat by optimizing feature map selection, following the method used in DepthSplat\cite{xu2024depthsplat}, to improve performance with more than two input views. To ensure a fair comparison, we additionally retrain pixelSplat on the DL3DV-10K dataset using configurations with 4 to 6 input views.
\subsubsection{Quantitative and Qualitative results}
Quantitative comparisons across all baselines are presented in Table~\ref{tab:dl3dv_n300}, with the number of input views ranging from 4 to 6. As a result, ProSplat consistently outperforms all baselines in all evaluation metrics, achieving particularly notable gains in PSNR and LPIPS. The significant improvement in LPIPS can be attributed to the use of the 2D priors from the diffusion model, which enhances the clarity in regions that suffer from severe blurring. Qualitative comparisons are illustrated in Figure~\ref{fig:comparison_grid}, where ProSplat clearly delivers superior visual quality. Except for MVSplat360, other methods exhibit prominent blur and structural degradation in challenging areas. Although MVSplat360 can effectively suppress blurring, it often introduces hallucinated artifacts that undermine 3D geometric consistency. In contrast, ProSplat mitigates these artifacts using a one-step diffusion model, which better preserves the target view structure compared to multi-step latent diffusion models\cite{rombach2022ldm}. In addition, the integration of DWEA guides feature alignment, enhancing 3D geometric consistency across views.

Because most scenes in the DL3DV-10K dataset include two-round camera scan trajectories, we additionally evaluate ProSplat using views from only the first scan round to assess its novel view synthesis performance. The results presented in Table~\ref{tab:dl3dv_n300} demonstrate that ProSplat exceeds all baselines in all metrics. 

\newlength{\imagewidth}
\setlength{\imagewidth}{\dimexpr (\textwidth - 3mm)/6}
\begin{figure*}[ht]
    \centering

    \makebox[\imagewidth]{\centering \scriptsize \textbf{GT}}%
    \makebox[\imagewidth]{\centering \scriptsize \textbf{pixelSplat\cite{pixelsplat}}}\hspace{0.5mm}%
    \makebox[\imagewidth]{\centering \scriptsize \textbf{MVSplat\cite{mvsplat}}}\hspace{0.5mm}%
    \makebox[\imagewidth]{\centering \scriptsize \textbf{DepthSplat\cite{xu2024depthsplat}}}\hspace{0.5mm}%
    \makebox[\imagewidth]{\centering \scriptsize \textbf{MVSplat360\cite{chen2024mvsplat360}}}\hspace{0.5mm}%
    \makebox[\imagewidth]{\centering \scriptsize \textbf{Ours}}\hspace{0.5mm}\\[1mm]

    \includegraphics[width=\imagewidth]{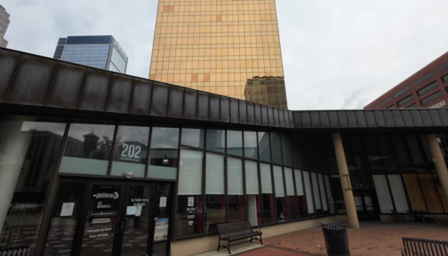}\hspace{0.5mm}%
    \includegraphics[width=\imagewidth]{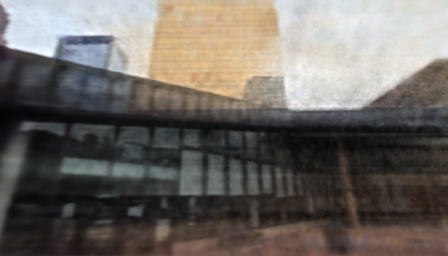}\hspace{0.5mm}%
    \includegraphics[width=\imagewidth]{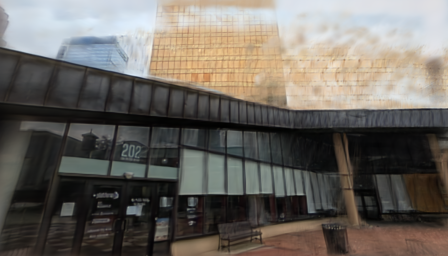}\hspace{0.5mm}%
    \includegraphics[width=\imagewidth]{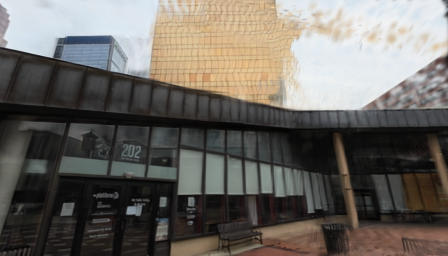}\hspace{0.5mm}%
    \includegraphics[width=\imagewidth]{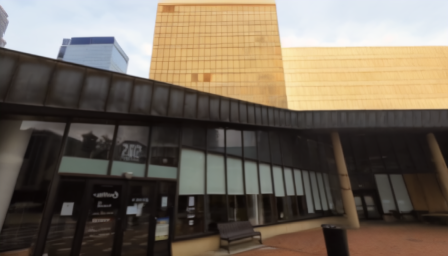}\hspace{0.5mm}%
    \includegraphics[width=\imagewidth]{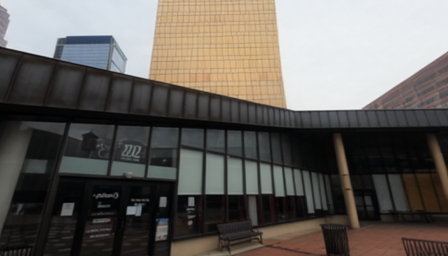} \\[0.5mm]

    \includegraphics[width=\imagewidth]{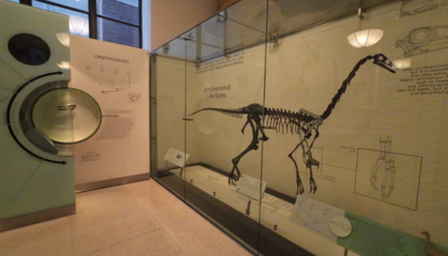}\hspace{0.5mm}%
    \includegraphics[width=\imagewidth]{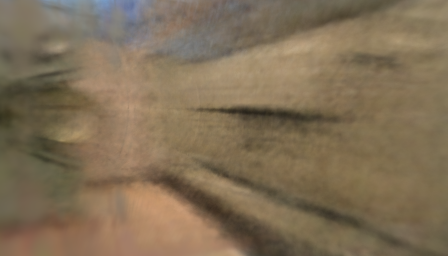}\hspace{0.5mm}%
    \includegraphics[width=\imagewidth]{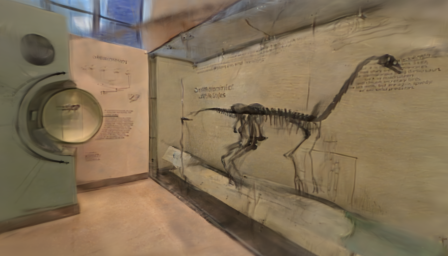}\hspace{0.5mm}%
    \includegraphics[width=\imagewidth]{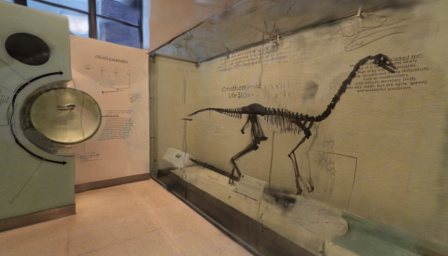}\hspace{0.5mm}%
    \includegraphics[width=\imagewidth]{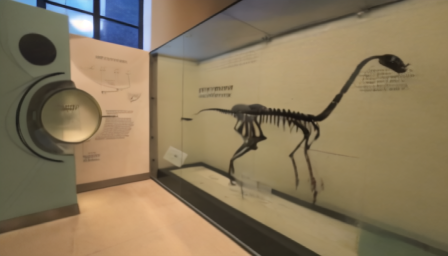}\hspace{0.5mm}%
    \includegraphics[width=\imagewidth]{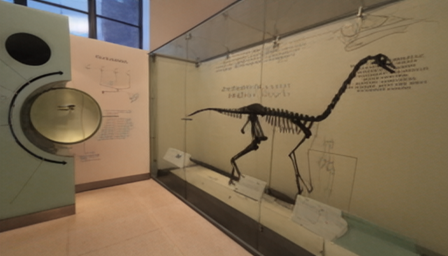} \\[0.5mm]

    \includegraphics[width=\imagewidth]{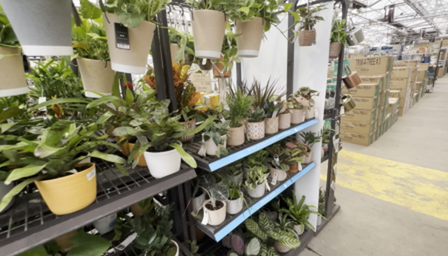}\hspace{0.5mm}%
    \includegraphics[width=\imagewidth]{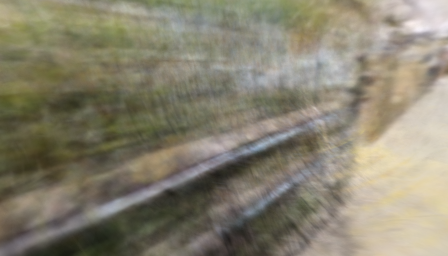}\hspace{0.5mm}%
    \includegraphics[width=\imagewidth]{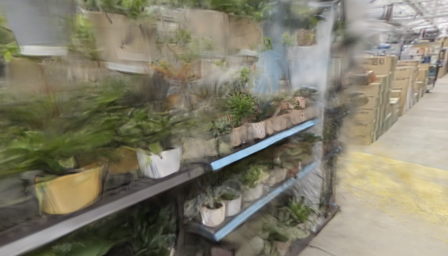}\hspace{0.5mm}%
    \includegraphics[width=\imagewidth]{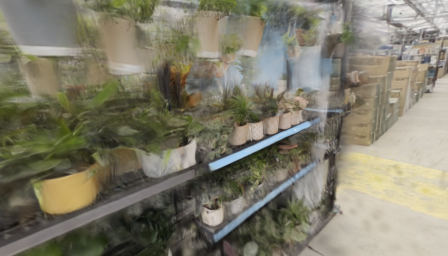}\hspace{0.5mm}%
    \includegraphics[width=\imagewidth]{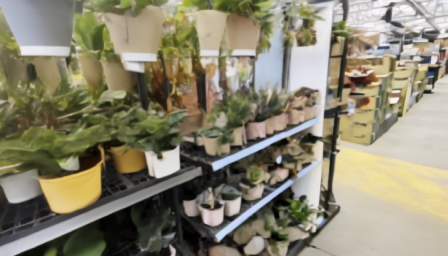}\hspace{0.5mm}%
    \includegraphics[width=\imagewidth]{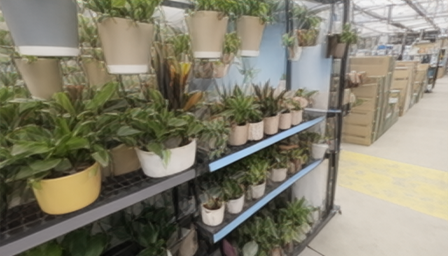} \\[0.5mm]

    \includegraphics[width=\imagewidth]{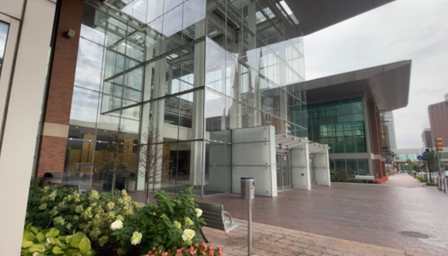}\hspace{0.5mm}%
    \includegraphics[width=\imagewidth]{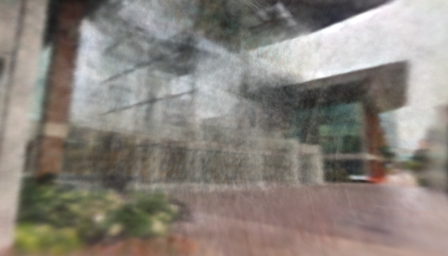}\hspace{0.5mm}%
    \includegraphics[width=\imagewidth]{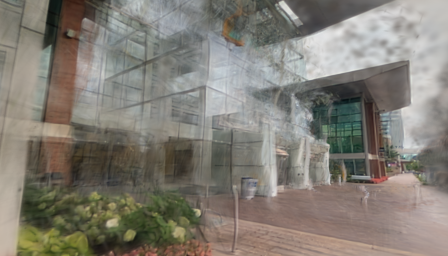}\hspace{0.5mm}%
    \includegraphics[width=\imagewidth]{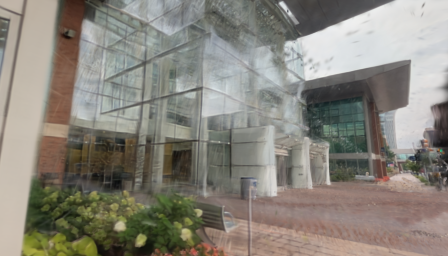}\hspace{0.5mm}%
    \includegraphics[width=\imagewidth]{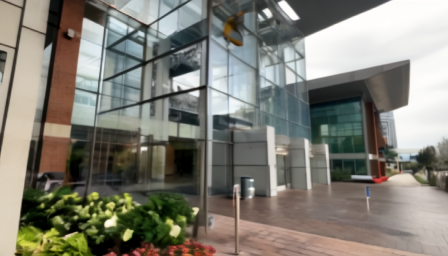}\hspace{0.5mm}%
    \includegraphics[width=\imagewidth]{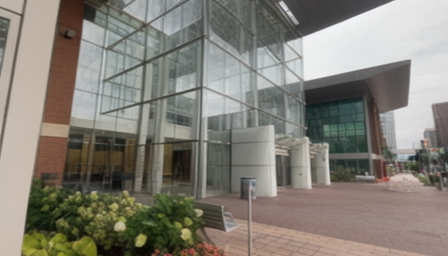} \\[0.5mm]
    \includegraphics[width=\imagewidth]{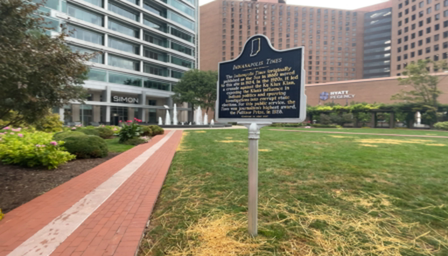}\hspace{0.5mm}%
    \includegraphics[width=\imagewidth]{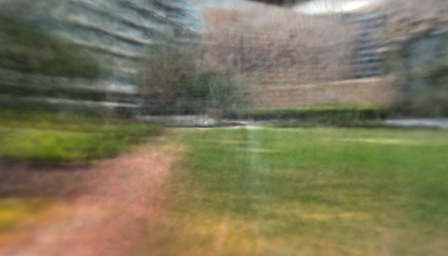}\hspace{0.5mm}%
    \includegraphics[width=\imagewidth]{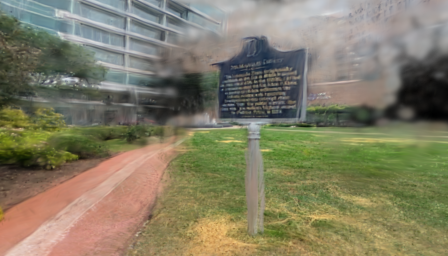}\hspace{0.5mm}%
    \includegraphics[width=\imagewidth]{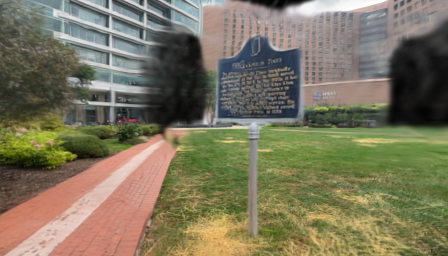}\hspace{0.5mm}%
    \includegraphics[width=\imagewidth]{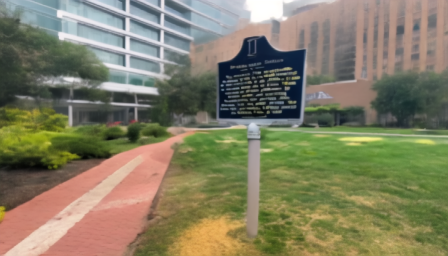}\hspace{0.5mm}%
    \includegraphics[width=\imagewidth]{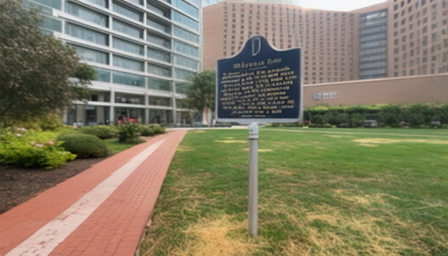} \\[0.5mm]
    \includegraphics[width=\imagewidth]{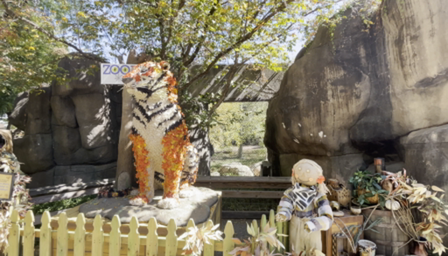}\hspace{0.5mm}%
    \includegraphics[width=\imagewidth]{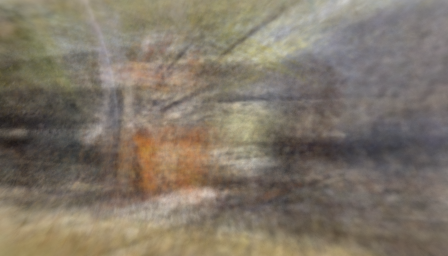}\hspace{0.5mm}%
    \includegraphics[width=\imagewidth]{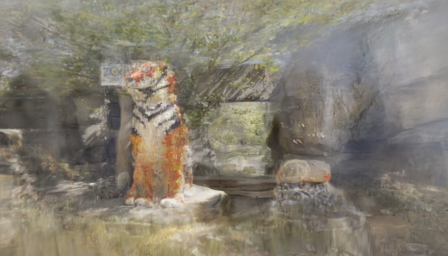}\hspace{0.5mm}%
    \includegraphics[width=\imagewidth]{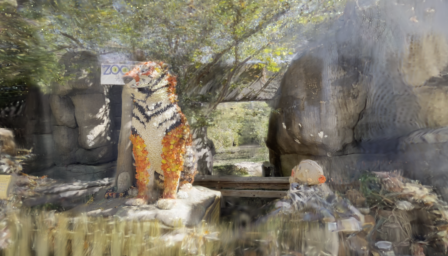}\hspace{0.5mm}%
    \includegraphics[width=\imagewidth]{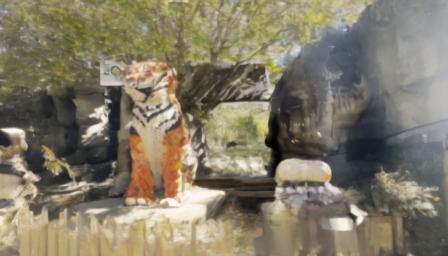}\hspace{0.5mm}%
    \includegraphics[width=\imagewidth]{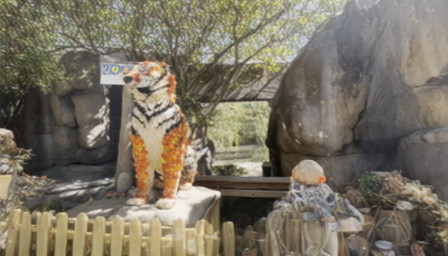} \\[0.5mm]

    \caption{\textbf{Qualitative comparisons on DL3DV-10K.} We test all baselines on the DL3DV-10K benchmark. ProSplat is able to restore blurry regions while demonstrating high perceptual quality for human perception. Although MVSplat360 can improve the quality of rendered views, it faces the challenge of generating excessive hallucination artifacts and often produces over-saturated results.}
    \label{fig:comparison_grid}
\end{figure*}
\begin{table}[t]
    \centering
    \caption{\textbf{Comparison with state-of-the-art methods on RealEstate10K with 2 input views.}} 
    \label{tab:re10k_2views}
    \renewcommand{\arraystretch}{1.2}
    \begin{tabular}{
        l
        ccc@{\hskip 20pt}
        ccc@{\hskip 20pt}
        ccc
    }
        \toprule
        \textbf{Method} & PSNR$\uparrow$ & SSIM$\uparrow$ & LPIPS$\downarrow$ & FID$\downarrow$ \\
        \midrule
        pixelSplat\cite{pixelsplat} 
        & 20.69 & 0.731 & 0.253 & 12.19\\
        
        MVSplat\cite{mvsplat} 
        & 20.95 & 0.744 & 0.233 & 9.64\\
        
        MVSplat360\cite{chen2024mvsplat360} 
        & 21.34 & 0.768 & 0.192 & 8.50\\
        NoPoSplat\cite{noposplat} 
        & 22.90 & 0.796 & 0.201 & 8.93\\           
        DepthSplat\cite{xu2024depthsplat} 
        & 22.94 & 0.800 & 0.190 & 6.93\\       
        \textbf{ProSplat} (Ours)  & \textbf{23.58} & \textbf{0.809} & \textbf{0.136} & \textbf{5.53}\\
        
        \bottomrule
    \end{tabular}
\end{table}
\begin{figure*}[t]
\centering
\includegraphics[width=\linewidth, trim=0cm 14.4cm 0cm 0cm, clip]{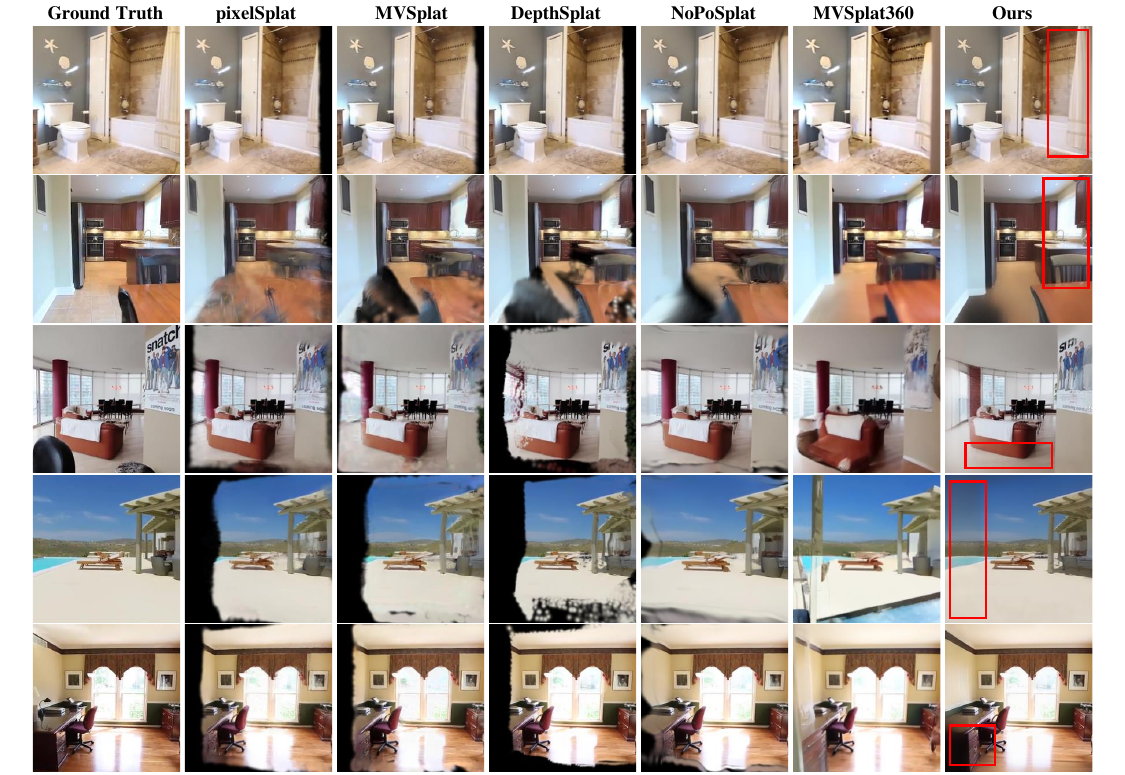}
\caption{\textbf{Qualitative comparisons of extrapolation performance on RealEstate10K dataset.} ProSplat generates plausible texture and color in extrapolated views of previously unobserved regions, whereas MVSplat360 tends to fill in content hallucination artifacts.}
\label{fig:comparison_img_re10k}
\end{figure*}

\subsection{Results on RealEstate10K Dataset}
\subsubsection{Baselines}
We adopt the same baselines as those evaluated on DL3DV-10K dataset, with the addition of NoPoSplat\cite{noposplat}, which classifies scenes by view overlap factors. Because NoPoSplat accepts only two or three input views, it is not evaluated on DL3DV-10K. For MVSplat360, we fine-tune the publicly released checkpoint on DL3DV-10K for 100 K steps and apply a post-process to mitigate over-saturation. For MVSplat, we retrain the model with the feature selection strategy proposed in DepthSplat \cite{xu2024depthsplat} to improve performance when more than two input views are provided.
\subsubsection{Quantitative and Qualitative results}
We first evaluate all methods in a wide-baseline setting using two input views, with the results reported in Table~\ref{tab:re10k_2views}. ProSplat achieves higher performance than all baselines in every metric. The improvements in PSNR and SSIM are relatively moderate, which can be attributed to the already high-quality renderings produced by the baselines. In contrast, LPIPS shows a significant increase, highlighting the perceptual advantages of ProSplat.

We then evaluate extrapolative novel view synthesis using the view configuration proposed by MVSplat360~\cite{chen2024mvsplat360}. The corresponding quantitative results are presented in Table~\ref{tab:re10k_extra}. ProSplat outperforms all baselines on every metric except FID, where MVSplat360 achieves a lower score. This outcome can be attributed to the more complex latent diffusion model in MVSplat360, which better aligns with the distribution of real images. However, a lower FID does not necessarily indicate better visual fidelity. As shown in the qualitative comparisons in Figure~\ref{fig:comparison_img_re10k}, MVSplat360 often introduces hallucinated artifacts due to its high generative capacity. In contrast, ProSplat produces plausible texture and color in previously incomplete regions, while effectively avoiding such artifacts and preserving 3D geometric consistency.  
\begin{table}[t]
    \centering
    \caption{\textbf{Comparison of extrapolation performance on RealEstate10K.}} 
    \label{tab:re10k_extra}
    \renewcommand{\arraystretch}{1.2}
    \begin{tabular}{
        l
        ccc@{\hskip 20pt}
        ccc@{\hskip 20pt}
        ccc
    }
        \toprule
        \textbf{Method} & PSNR$\uparrow$ & SSIM$\uparrow$ & LPIPS$\downarrow$ & FID$\downarrow$ \\
        \midrule
        pixelSplat\cite{pixelsplat} 
        & 21.84 & 0.777 & 0.216 & 5.78\\
        
        MVSplat\cite{mvsplat} 
        & 23.04 & 0.812 & 0.185 & 3.83\\
        
        MVSplat360\cite{chen2024mvsplat360} 
        & 23.16 & 0.810 & 0.176 & \textbf{1.79}\\   

        DepthSplat\cite{xu2024depthsplat} 
        & 20.53 & 0.799 & 0.204 & 6.93\\       
        NoPoSplat\cite{chen2024mvsplat360} 
        & 23.07 & 0.810 & 0.191 & 5.66\\ 
        \textbf{ProSplat} (Ours)  & \textbf{23.97} & \textbf{0.839} & \textbf{0.158} & 3.65\\
        \bottomrule
    \end{tabular}
\end{table}

\begin{table*}[t]
  \centering
    \caption{\textbf{Model components of improvement mode.} The baseline version of our improvement model consists of a VAE and a U-Net, excluding the denoising process commonly used in diffusion models. MORI and DWEA denote Maximum Overlap Reference view Injection and Distance-Weighted Epipolar Attention, respectively.} 
  \label{tab:ablation}
  \renewcommand{\arraystretch}{1.2}
  \setlength{\tabcolsep}{10pt}
  \begin{tabular}{cccc ccc}
    \toprule
     \multicolumn{4}{c}{\textbf{Components}} 
    & \multirow{2}{*}{\hspace{5pt}PSNR↑\hspace{5pt}} 
    & \multirow{2}{*}{\hspace{5pt}SSIM↑\hspace{5pt}} 
    & \multirow{2}{*}{\hspace{5pt}LPIPS↓\hspace{5pt}}  \\
    \cmidrule(lr){1-4}
    Improvement Model
    & Diffusion Model 
    & MORI + DWEA
    & Joint Training \\
    \midrule
     \ding{55} & \ding{55} & \ding{55} & \ding{55} & 17.87 & 0.559 & 0.371 \\
     \checkmark & \ding{55} & \ding{55} & \ding{55}  & 18.39 & 0.563 & 0.344  \\
     \checkmark & \checkmark & \ding{55} & \ding{55} & 18.74 & 0.573 & 0.315  \\
     \checkmark & \checkmark & \checkmark & \ding{55} & 18.84 & 0.580 & 0.313  \\
     \checkmark & \checkmark & \checkmark & \checkmark & \textbf{19.24} & \textbf{0.590} & \textbf{0.302} \\
    \bottomrule
  \end{tabular}
\end{table*}
\subsection{Ablations and Analysis}
We start by evaluating ProSplat without the improvement model. Subsequently, we incrementally add components of the improvement model,  including the diffusion model, MORI, and DWEA. The diffusion model here refers to the denoising process, excluding VAE and U-Net, which are already part of the baseline improvement model. Finally, all components are integrated and jointly optimized using our proposed joint training strategy.
We evaluate the effectiveness of each configuration using PSNR, SSIM, and LPIPS metrics. All ablation experiments are conducted on the DL3DV-10K dataset using 6 input views.

As shown in Table~\ref{tab:ablation}, the baseline improvement model already provides a significant performance gain. In the remainder of this section, we analyze the contribution of each component associated with the improvement model. 

\subsubsection{Effects of diffusion model}
\begin{figure}[t]
\centering
\includegraphics[width=\linewidth, trim=0cm 18.1cm 0cm 0cm, clip]{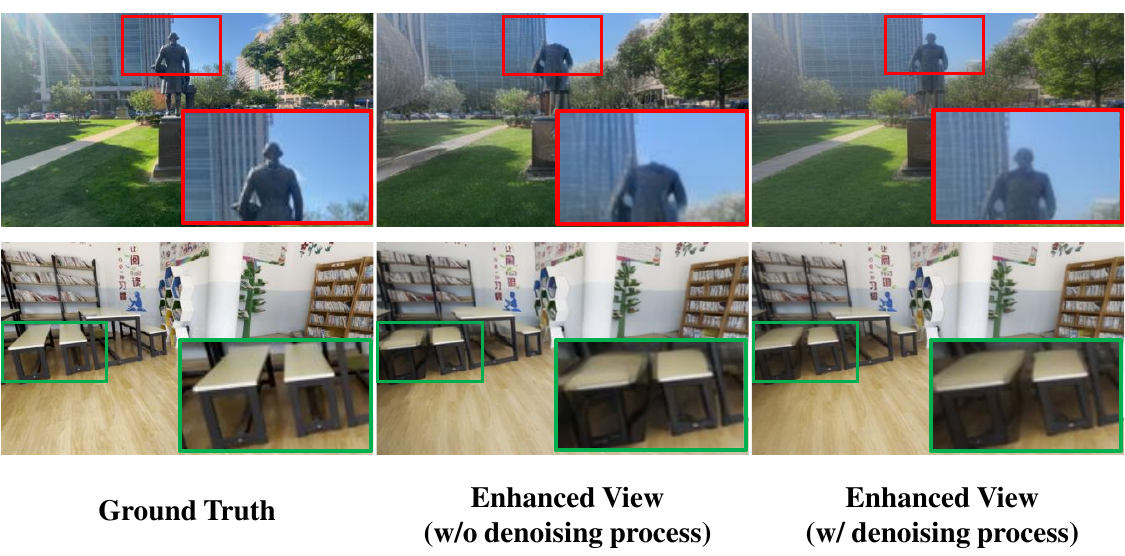}
\caption{\textbf{Effect of denoising process.} Without the denoising process, ProSplat has difficulty identifying object boundaries in regions with severe blurring. In the enhanced view \emph{without} denoising (center image, bottom row), the boundaries of the two desks are indistinct, 
whereas the corresponding view \emph{with} denoising (right image, bottom row) exhibits sharply defined desk contours.}
\label{fig:comparison_abl}
\end{figure}

As shown in Figure~\ref{fig:comparison_abl}, without the denoising process of the diffusion model, the model behaves similarly to a U-Net operating in latent space. 
Although it can improve the quality of the rendered images, it struggles to restore clarity in regions with severe blurring. In contrast, with the inclusion of the diffusion model, ProSplat more effectively enhances rendered images, even in unoccupied regions.

\subsubsection{Effects of MORI and DWEA}
DWEA is applied to fuse the bottleneck features of the reference and target views, making it intrinsically linked to the MORI strategy. Accordingly, we perform ablation studies by treating MORI and DWEA as a combined component.
Quantitative evaluations shown in Table~\ref{tab:ablation} indicate consistent improvements in all metrics, although the gains are relatively modest. This can be attributed to the fact that the diffusion model is pre-trained on a large-scale dataset, enabling it to infer relevant regions in some scenes using strong 2D priors.
Nevertheless, integrating MORI and DWEA facilitates a more accurate recovery of texture and color, while also enhancing 3D geometric consistency. 

\begin{table}[t]
  \centering
  \renewcommand{\arraystretch}{1.2}
  \caption{\textbf{Results of Substituting 3DGS Generator with MVSplat*.} MVSplat* serves as the 3DGS generator of MVSplat360.}
  \label{tab:backbone_test}
  \begin{tabular}{lcccc}
    \toprule
    \textbf{Method} & PSNR↑ & SSIM↑ & LPIPS↓ & FID↓ \\
    \midrule
    MVSplat*     & 16.77 & 0.485 & 0.456 & 62.51 \\
    MVSplat360   & 17.09 & \textbf{0.514} & 0.418 & \textbf{16.54} \\
    \textbf{ProSplat} (Ours)         & \textbf{17.40} & 0.503 & \textbf{0.383} & 20.45 \\
    \bottomrule
  \end{tabular}
\end{table}
\subsubsection{Joint Training}
As shown in Table~\ref{tab:ablation}, the model trained with joint optimization consistently outperforms the independently trained counterpart across all evaluation metrics. This performance gap can be attributed to the better distributional alignment between the 3DGS generator and the improvement model when they are optimized end-to-end. These results highlight the importance of joint optimization in harmonizing intermediate representations and ensuring consistent gradient flow across network components.
\subsubsection{Substituting 3DGS Generator}
To evaluate the robustness of the improvement model, we replace the original 3DGS generator with the generator used in MVSplat360, which is built on MVSplat\cite{mvsplat}. Unlike our original generator, the MVSplat360 generator is adapted to align with the input data distribution expected by the latent diffusion model\cite{rombach2022ldm}.
All evaluation metrics are presented in Table~\ref{tab:backbone_test}, based on experiments conducted on the DL3DV-10K dataset using six input views. Despite the mismatch in data distributions between the new generator and our improvement model, ProSplat still achieves high PSNR and LPIPS scores, demonstrating strong robustness under varying input conditions.
\section{Conclusion}
\label{sec:conclusion}
We introduce ProSplat, an efficient feed-forward framework for novel view synthesis from wide-baseline sparse views. By incorporating a carefully designed improvement model at the rendering stage, ProSplat achieves SOTA performance on two large-scale 3D reconstruction benchmarks: DL3DV-10K and RealEstate10K. Extensive experiments demonstrate that ProSplat effectively reconstructs blurred or missing regions and delivers high visual fidelity across challenging view synthesis scenarios. 
This performance benefits from the rich 2D priors embedded in the pre-trained diffusion model. 
Furthermore, the integration of MORI and DWEA enhances texture and color fidelity while preserving 3D geometric consistency. 
In contrast to MVSplat360, which relies on a multi-step latent diffusion pipeline in the feature space, ProSplat significantly reduces hallucination artifacts. In addition, ProSplat outperforms other recent feed-forward SOTA methods in both quantitative metrics and qualitative evaluations, demonstrating its robustness and generalizability across diverse scenes.

In summary, ProSplat offers a practical and high-quality solution for wide-baseline sparse-view novel view synthesis, with strong potential for real-world applications in immersive media, virtual reality, and 3D reconstruction.

\section{Limitations} 
\label{sec:limitation}
Despite these advantages, the single-step diffusion model has limited generative capacity and may produce flat or over-smoothed textures in viewpoints with minimal visual coverage. Moreover, because the view improvement is performed at the rendering stage, achieving real-time inference remains a challenge. Future work will explore heterogeneous acceleration strategies, such as integration of FPGA and GPU resources, to enable real-time processing. These advances could unlock high-performance, sparse-view 3D reconstruction for practical applications.

\bibliographystyle{IEEEtran}
\bibliography{TCSVT}

\begin{thebibliography}{10}
\providecommand{\url}[1]{#1}
\csname url@samestyle\endcsname
\providecommand{\newblock}{\relax}
\providecommand{\bibinfo}[2]{#2}
\providecommand{\BIBentrySTDinterwordspacing}{\spaceskip=0pt\relax}
\providecommand{\BIBentryALTinterwordstretchfactor}{4}
\providecommand{\BIBentryALTinterwordspacing}{\spaceskip=\fontdimen2\font plus
\BIBentryALTinterwordstretchfactor\fontdimen3\font minus \fontdimen4\font\relax}
\providecommand{\BIBforeignlanguage}[2]{{%
\expandafter\ifx\csname l@#1\endcsname\relax
\typeout{** WARNING: IEEEtran.bst: No hyphenation pattern has been}%
\typeout{** loaded for the language `#1'. Using the pattern for}%
\typeout{** the default language instead.}%
\else
\language=\csname l@#1\endcsname
\fi
#2}}
\providecommand{\BIBdecl}{\relax}
\BIBdecl

\bibitem{gao2021dynamic}
C.~Gao, A.~Saraf, J.~Kopf, and J.-B. Huang, ``Dynamic view synthesis from dynamic monocular video,'' in \emph{Proceedings of the IEEE/CVF International Conference on Computer Vision}, 2021, pp. 5712--5721.

\bibitem{attal2020matryodshka}
B.~Attal, S.~Ling, A.~Gokaslan, C.~Richardt, and J.~Tompkin, ``Matryodshka: Real-time 6dof video view synthesis using multi-sphere images,'' in \emph{European Conference on Computer Vision}.\hskip 1em plus 0.5em minus 0.4em\relax Springer, 2020, pp. 441--459.

\bibitem{jin2025rendering}
X.~Jin, Y.~Fang, M.~Frosi, J.~Ge, J.~Xiao, and M.~Matteucci, ``Rendering anywhere you see: Renderability field-guided gaussian splatting,'' \emph{arXiv preprint arXiv:2504.19261}, 2025.

\bibitem{mildenhall2021nerf}
B.~Mildenhall, P.~P. Srinivasan, M.~Tancik, J.~T. Barron, R.~Ramamoorthi, and R.~Ng, ``Nerf: Representing scenes as neural radiance fields for view synthesis,'' \emph{Communications of the ACM}, vol.~65, no.~1, pp. 99--106, 2021.

\bibitem{muller2022instant}
T.~M{\"u}ller, A.~Evans, C.~Schied, and A.~Keller, ``Instant neural graphics primitives with a multiresolution hash encoding,'' \emph{ACM Transactions on Graphics (TOG)}, vol.~41, no.~4, pp. 1--15, 2022.

\bibitem{fridovich2022plenoxels}
S.~Fridovich-Keil, A.~Yu, M.~Tancik, Q.~Chen, B.~Recht, and A.~Kanazawa, ``Plenoxels: Radiance fields without neural networks,'' in \emph{Proceedings of the IEEE/CVF Conference on Computer Vision and Pattern Recognition}, 2022, pp. 5501--5510.

\bibitem{kerbl20233d3dgs}
B.~Kerbl, G.~Kopanas, T.~Leimk{\"u}hler, and G.~Drettakis, ``3d gaussian splatting for real-time radiance field rendering.'' \emph{ACM Transactions on Graphics (TOG)}, vol.~42, no.~4, pp. 139--1, 2023.

\bibitem{kopanas2022neural}
G.~Kopanas, T.~Leimk{\"u}hler, G.~Rainer, C.~Jambon, and G.~Drettakis, ``Neural point catacaustics for novel-view synthesis of reflections,'' \emph{ACM Transactions on Graphics (TOG)}, vol.~41, no.~6, pp. 1--15, 2022.

\bibitem{pixelsplat}
D.~Charatan, S.~L. Li, A.~Tagliasacchi, and V.~Sitzmann, ``pixelsplat: 3d gaussian splats from image pairs for scalable generalizable 3d reconstruction,'' in \emph{Proceedings of the IEEE/CVF Conference on Computer Vision and Pattern Recognition}, 2024, pp. 19\,457--19\,467.

\bibitem{mvsplat}
Y.~Chen, H.~Xu, C.~Zheng, B.~Zhuang, M.~Pollefeys, A.~Geiger, T.-J. Cham, and J.~Cai, ``Mvsplat: Efficient 3d gaussian splatting from sparse multi-view images,'' in \emph{European Conference on Computer Vision}.\hskip 1em plus 0.5em minus 0.4em\relax Springer, 2024, pp. 370--386.

\bibitem{xu2024depthsplat}
H.~Xu, S.~Peng, F.~Wang, H.~Blum, D.~Barath, A.~Geiger, and M.~Pollefeys, ``Depthsplat: Connecting gaussian splatting and depth,'' in \emph{Proceedings of the IEEE/CVF Conference on Computer Vision and Pattern Recognition}, 2025.

\bibitem{zhang2024gaussian}
S.~Zhang, X.~Fei, F.~Liu, H.~Song, and Y.~Duan, ``Gaussian graph network: Learning efficient and generalizable gaussian representations from multi-view images,'' \emph{Advances in Neural Information Processing Systems}, vol.~37, pp. 50\,361--50\,380, 2024.

\bibitem{yu2021pixelnerf}
A.~Yu, V.~Ye, M.~Tancik, and A.~Kanazawa, ``pixelnerf: Neural radiance fields from one or few images,'' in \emph{Proceedings of the IEEE/CVF Conference on Computer Vision and Pattern Recognition}, 2021, pp. 4578--4587.

\bibitem{yu2024viewcrafter}
W.~Yu, J.~Xing, L.~Yuan, W.~Hu, X.~Li, Z.~Huang, X.~Gao, T.-T. Wong, Y.~Shan, and Y.~Tian, ``Viewcrafter: Taming video diffusion models for high-fidelity novel view synthesis,'' \emph{arXiv preprint arXiv:2409.02048}, 2024.

\bibitem{liu20243dgs}
X.~Liu, C.~Zhou, and S.~Huang, ``3dgs-enhancer: Enhancing unbounded 3d gaussian splatting with view-consistent 2d diffusion priors,'' \emph{Advances in Neural Information Processing Systems}, vol.~37, pp. 133\,305--133\,327, 2024.

\bibitem{wu2025difix3d}
J.~Z. Wu, Y.~Zhang, H.~Turki, X.~Ren, J.~Gao, M.~Z. Shou, S.~Fidler, Z.~Gojcic, and H.~Ling, ``Difix3d+: Improving 3d reconstructions with single-step diffusion models,'' in \emph{Proceedings of the IEEE/CVF Conference on Computer Vision and Pattern Recognition}, 2025.

\bibitem{blattmann2023stable}
A.~Blattmann, T.~Dockhorn, S.~Kulal, D.~Mendelevitch, M.~Kilian, D.~Lorenz, Y.~Levi, Z.~English, V.~Voleti, A.~Letts \emph{et~al.}, ``Stable video diffusion: Scaling latent video diffusion models to large datasets,'' \emph{arXiv preprint arXiv:2311.15127}, 2023.

\bibitem{xing2024dynamicrafter}
J.~Xing, M.~Xia, Y.~Zhang, H.~Chen, W.~Yu, H.~Liu, G.~Liu, X.~Wang, Y.~Shan, and T.-T. Wong, ``Dynamicrafter: Animating open-domain images with video diffusion priors,'' in \emph{European Conference on Computer Vision}.\hskip 1em plus 0.5em minus 0.4em\relax Springer, 2024, pp. 399--417.

\bibitem{rombach2022ldm}
R.~Rombach, A.~Blattmann, D.~Lorenz, P.~Esser, and B.~Ommer, ``High-resolution image synthesis with latent diffusion models,'' in \emph{Proceedings of the IEEE/CVF Conference on Computer Vision and Pattern Recognition}, 2022, pp. 10\,684--10\,695.

\bibitem{chen2024mvsplat360}
Y.~Chen, C.~Zheng, H.~Xu, B.~Zhuang, A.~Vedaldi, T.-J. Cham, and J.~Cai, ``Mvsplat360: Feed-forward 360 scene synthesis from sparse views,'' in \emph{Advances in Neural Information Processing Systems (NeurIPS)}, 2024.

\bibitem{parmar2024img2img}
G.~Parmar, T.~Park, S.~Narasimhan, and J.-Y. Zhu, ``One-step image translation with text-to-image models,'' \emph{arXiv preprint arXiv:2403.12036}, 2024.

\bibitem{yin2024one}
T.~Yin, M.~Gharbi, R.~Zhang, E.~Shechtman, F.~Durand, W.~T. Freeman, and T.~Park, ``One-step diffusion with distribution matching distillation,'' in \emph{Proceedings of the IEEE/CVF Conference on Computer Vision and Pattern Recognition}, 2024, pp. 6613--6623.

\bibitem{sauer2024adversarial_sd-turbo}
A.~Sauer, D.~Lorenz, A.~Blattmann, and R.~Rombach, ``Adversarial diffusion distillation,'' in \emph{European Conference on Computer Vision}.\hskip 1em plus 0.5em minus 0.4em\relax Springer, 2024, pp. 87--103.

\bibitem{hu2022lora}
E.~J. Hu, Y.~Shen, P.~Wallis, Z.~Allen-Zhu, Y.~Li, S.~Wang, L.~Wang, W.~Chen \emph{et~al.}, ``Lora: Low-rank adaptation of large language models.'' \emph{ICLR}, vol.~1, no.~2, p.~3, 2022.

\bibitem{kingma2013auto}
D.~P. Kingma, M.~Welling \emph{et~al.}, ``Auto-encoding variational bayes,'' 2013.

\bibitem{ronneberger2015u}
O.~Ronneberger, P.~Fischer, and T.~Brox, ``U-net: Convolutional networks for biomedical image segmentation,'' in \emph{Medical image computing and computer-assisted intervention--MICCAI 2015: 18th international conference, Munich, Germany, October 5-9, 2015, proceedings, part III 18}.\hskip 1em plus 0.5em minus 0.4em\relax Springer, 2015, pp. 234--241.

\bibitem{ling2024dl3dv}
L.~Ling, Y.~Sheng, Z.~Tu, W.~Zhao, C.~Xin, K.~Wan, L.~Yu, Q.~Guo, Z.~Yu, Y.~Lu \emph{et~al.}, ``Dl3dv-10k: A large-scale scene dataset for deep learning-based 3d vision,'' in \emph{Proceedings of the IEEE/CVF Conference on Computer Vision and Pattern Recognition}, 2024, pp. 22\,160--22\,169.

\bibitem{zhou2018re10k}
T.~Zhou, R.~Tucker, J.~Flynn, G.~Fyffe, and N.~Snavely, ``Stereo magnification: Learning view synthesis using multiplane images,'' in \emph{SIGGRAPH}, 2018.

\bibitem{wang2004ssim}
Z.~Wang, A.~C. Bovik, H.~R. Sheikh, and E.~P. Simoncelli, ``Image quality assessment: from error visibility to structural similarity,'' \emph{IEEE transactions on image processing}, vol.~13, no.~4, pp. 600--612, 2004.

\bibitem{zhang2018lpips}
R.~Zhang, P.~Isola, A.~A. Efros, E.~Shechtman, and O.~Wang, ``The unreasonable effectiveness of deep features as a perceptual metric,'' in \emph{Proceedings of the IEEE/CVF Conference on Computer Vision and Pattern Recognition}, 2018, pp. 586--595.

\bibitem{chollet2017xception}
F.~Chollet, ``Xception: Deep learning with depthwise separable convolutions,'' in \emph{Proceedings of the IEEE/CVF Conference on Computer Vision and Pattern Recognition}, 2017, pp. 1251--1258.

\bibitem{chen2023nvs1}
S.~E. Chen and L.~Williams, ``View interpolation for image synthesis,'' in \emph{Seminal Graphics Papers: Pushing the Boundaries, Volume 2}, 2023, pp. 423--432.

\bibitem{gortler2023nvs2}
S.~J. Gortler, R.~Grzeszczuk, R.~Szeliski, and M.~F. Cohen, ``The lumigraph,'' in \emph{Seminal Graphics Papers: Pushing the Boundaries, Volume 2}, 2023, pp. 453--464.

\bibitem{flynn2016deepstereonvs}
J.~Flynn, I.~Neulander, J.~Philbin, and N.~Snavely, ``Deepstereo: Learning to predict new views from the world's imagery,'' in \emph{Proceedings of the IEEE/CVF Conference on Computer Vision and Pattern Recognition}, 2016, pp. 5515--5524.

\bibitem{kalantari2016learningnvs}
N.~K. Kalantari, T.-C. Wang, and R.~Ramamoorthi, ``Learning-based view synthesis for light field cameras,'' \emph{ACM Transactions on Graphics (TOG)}, vol.~35, no.~6, Dec. 2016.

\bibitem{wang2024dust3r}
S.~Wang, V.~Leroy, Y.~Cabon, B.~Chidlovskii, and J.~Revaud, ``Dust3r: Geometric 3d vision made easy,'' in \emph{Proceedings of the IEEE/CVF Conference on Computer Vision and Pattern Recognition}, 2024, pp. 20\,697--20\,709.

\bibitem{niemeyer2022regnerf}
M.~Niemeyer, J.~T. Barron, B.~Mildenhall, M.~S. Sajjadi, A.~Geiger, and N.~Radwan, ``Regnerf: Regularizing neural radiance fields for view synthesis from sparse inputs,'' in \emph{Proceedings of the IEEE/CVF Conference on Computer Vision and Pattern Recognition}, 2022, pp. 5480--5490.

\bibitem{somraj2023simplenerf}
N.~Somraj, A.~Karanayil, and R.~Soundararajan, ``Simplenerf: Regularizing sparse input neural radiance fields with simpler solutions,'' in \emph{SIGGRAPH Asia 2023 Conference Papers}, 2023, pp. 1--11.

\bibitem{deng2022depthnerf}
K.~Deng, A.~Liu, J.-Y. Zhu, and D.~Ramanan, ``Depth-supervised nerf: Fewer views and faster training for free,'' in \emph{Proceedings of the IEEE/CVF Conference on Computer Vision and Pattern Recognition}, 2022, pp. 12\,882--12\,891.

\bibitem{zhu2024fsgs}
Z.~Zhu, Z.~Fan, Y.~Jiang, and Z.~Wang, ``Fsgs: Real-time few-shot view synthesis using gaussian splatting,'' in \emph{European conference on computer vision}.\hskip 1em plus 0.5em minus 0.4em\relax Springer, 2024, pp. 145--163.

\bibitem{yu2022monosdnerf}
Z.~Yu, S.~Peng, M.~Niemeyer, T.~Sattler, and A.~Geiger, ``Monosdf: Exploring monocular geometric cues for neural implicit surface reconstruction,'' \emph{Advances in neural information processing systems}, vol.~35, pp. 25\,018--25\,032, 2022.

\bibitem{park2025dropgaussian}
H.~Park, G.~Ryu, and W.~Kim, ``Dropgaussian: Structural regularization for sparse-view gaussian splatting,'' \emph{arXiv preprint arXiv:2504.00773}, 2025.

\bibitem{ma2025novel}
X.~Ma, J.~Zhang, P.~Lu, S.~Xu, and C.~Pan, ``Novel view synthesis under large-deviation viewpoint for autonomous driving,'' in \emph{Proceedings of the AAAI Conference on Artificial Intelligence}, vol.~39, no.~6, 2025, pp. 6000--6008.

\bibitem{zhang2024gs}
K.~Zhang, S.~Bi, H.~Tan, Y.~Xiangli, N.~Zhao, K.~Sunkavalli, and Z.~Xu, ``Gs-lrm: Large reconstruction model for 3d gaussian splatting,'' in \emph{European Conference on Computer Vision}.\hskip 1em plus 0.5em minus 0.4em\relax Springer, 2024, pp. 1--19.

\bibitem{tang2024lgm}
J.~Tang, Z.~Chen, X.~Chen, T.~Wang, G.~Zeng, and Z.~Liu, ``Lgm: Large multi-view gaussian model for high-resolution 3d content creation,'' in \emph{European Conference on Computer Vision}.\hskip 1em plus 0.5em minus 0.4em\relax Springer, 2024, pp. 1--18.

\bibitem{szymanowicz2024splatter}
S.~Szymanowicz, C.~Rupprecht, and A.~Vedaldi, ``Splatter image: Ultra-fast single-view 3d reconstruction,'' in \emph{Proceedings of the IEEE/CVF Conference on Computer Vision and Pattern Recognition}, 2024, pp. 10\,208--10\,217.

\bibitem{zheng2024gps}
S.~Zheng, B.~Zhou, R.~Shao, B.~Liu, S.~Zhang, L.~Nie, and Y.~Liu, ``Gps-gaussian: Generalizable pixel-wise 3d gaussian splatting for real-time human novel view synthesis,'' in \emph{Proceedings of the IEEE/CVF Conference on Computer Vision and Pattern Recognition}, 2024, pp. 19\,680--19\,690.

\bibitem{zhou2025gps}
B.~Zhou, S.~Zheng, H.~Tu, R.~Shao, B.~Liu, S.~Zhang, L.~Nie, and Y.~Liu, ``Gps-gaussian+: Generalizable pixel-wise 3d gaussian splatting for real-time human-scene rendering from sparse views,'' \emph{IEEE Transactions on Pattern Analysis and Machine Intelligence}, 2025.

\bibitem{noposplat}
B.~Ye, S.~Liu, H.~Xu, X.~Li, M.~Pollefeys, M.-H. Yang, and S.~Peng, ``No pose, no problem: Surprisingly simple 3d gaussian splats from sparse unposed images,'' \emph{arXiv preprint arXiv:2410.24207}, 2024.

\bibitem{wewer2024latentsplat}
C.~Wewer, K.~Raj, E.~Ilg, B.~Schiele, and J.~E. Lenssen, ``latentsplat: Autoencoding variational gaussians for fast generalizable 3d reconstruction,'' in \emph{European Conference on Computer Vision}.\hskip 1em plus 0.5em minus 0.4em\relax Springer, 2024, pp. 456--473.

\bibitem{liu2023zero}
R.~Liu, R.~Wu, B.~Van~Hoorick, P.~Tokmakov, S.~Zakharov, and C.~Vondrick, ``Zero-1-to-3: Zero-shot one image to 3d object,'' in \emph{Proceedings of the IEEE/CVF International Conference on Computer Vision}, 2023, pp. 9298--9309.

\bibitem{shi2023zero123++}
R.~Shi, H.~Chen, Z.~Zhang, M.~Liu, C.~Xu, X.~Wei, L.~Chen, C.~Zeng, and H.~Su, ``Zero123++: a single image to consistent multi-view diffusion base model,'' \emph{arXiv preprint arXiv:2310.15110}, 2023.

\bibitem{shi2023mvdream}
Y.~Shi, P.~Wang, J.~Ye, M.~Long, K.~Li, and X.~Yang, ``Mvdream: Multi-view diffusion for 3d generation,'' \emph{arXiv preprint arXiv:2308.16512}, 2023.

\bibitem{huang2024epidiff}
Z.~Huang, H.~Wen, J.~Dong, Y.~Wang, Y.~Li, X.~Chen, Y.-P. Cao, D.~Liang, Y.~Qiao, B.~Dai \emph{et~al.}, ``Epidiff: Enhancing multi-view synthesis via localized epipolar-constrained diffusion,'' in \emph{Proceedings of the IEEE/CVF Conference on Computer Vision and Pattern Recognition}, 2024, pp. 9784--9794.

\bibitem{zhou2023sparsefusion}
Z.~Zhou and S.~Tulsiani, ``Sparsefusion: Distilling view-conditioned diffusion for 3d reconstruction,'' in \emph{Proceedings of the IEEE/CVF Conference on Computer Vision and Pattern Recognition}, 2023, pp. 12\,588--12\,597.

\bibitem{wu2025genfusion}
S.~Wu, C.~Xu, B.~Huang, A.~Geiger, and A.~Chen, ``Genfusion: Closing the loop between reconstruction and generation via videos,'' \emph{arXiv preprint arXiv:2503.21219}, 2025.

\bibitem{gu2023nerfdiff}
J.~Gu, A.~Trevithick, K.-E. Lin, J.~M. Susskind, C.~Theobalt, L.~Liu, and R.~Ramamoorthi, ``Nerfdiff: Single-image view synthesis with nerf-guided distillation from 3d-aware diffusion,'' in \emph{International Conference on Machine Learning}.\hskip 1em plus 0.5em minus 0.4em\relax PMLR, 2023, pp. 11\,808--11\,826.

\bibitem{chan2023generative}
E.~R. Chan, K.~Nagano, M.~A. Chan, A.~W. Bergman, J.~J. Park, A.~Levy, M.~Aittala, S.~De~Mello, T.~Karras, and G.~Wetzstein, ``Generative novel view synthesis with 3d-aware diffusion models,'' in \emph{Proceedings of the IEEE/CVF International Conference on Computer Vision}, 2023, pp. 4217--4229.

\bibitem{zwicker2001surface}
M.~Zwicker, H.~Pfister, J.~Van~Baar, and M.~Gross, ``Surface splatting,'' in \emph{Proceedings of the 28th annual conference on Computer graphics and interactive techniques}, 2001, pp. 371--378.

\bibitem{liu2021swin}
Z.~Liu, Y.~Lin, Y.~Cao, H.~Hu, Y.~Wei, Z.~Zhang, S.~Lin, and B.~Guo, ``Swin transformer: Hierarchical vision transformer using shifted windows,'' in \emph{Proceedings of the IEEE/CVF International Conference on Computer Vision}, 2021, pp. 10\,012--10\,022.

\bibitem{vaswani2017attention}
A.~Vaswani, N.~Shazeer, N.~Parmar, J.~Uszkoreit, L.~Jones, A.~N. Gomez, {\L}.~Kaiser, and I.~Polosukhin, ``Attention is all you need,'' \emph{Advances in neural information processing systems}, vol.~30, 2017.

\bibitem{paszke2019pytorch}
A.~Paszke, ``Pytorch: An imperative style, high-performance deep learning library,'' \emph{arXiv preprint arXiv:1912.01703}, 2019.

\bibitem{edstedt2024roma}
J.~Edstedt, Q.~Sun, G.~B{\"o}kman, M.~Wadenb{\"a}ck, and M.~Felsberg, ``Roma: Robust dense feature matching,'' in \emph{Proceedings of the IEEE/CVF Conference on Computer Vision and Pattern Recognition}, 2024, pp. 19\,790--19\,800.

\bibitem{heusel2017fid}
M.~Heusel, H.~Ramsauer, T.~Unterthiner, B.~Nessler, and S.~Hochreiter, ``Gans trained by a two time-scale update rule converge to a local nash equilibrium,'' \emph{Advances in neural information processing systems}, vol.~30, 2017.

\end{thebibliography}

\end{document}